\documentclass[graybox,envcountchap]{svmult}

\usepackage[T1]{fontenc}
\usepackage{mathptmx}        
\usepackage{helvet}          
\usepackage{courier}         
\usepackage{textcomp}

\usepackage{makeidx}         
\usepackage{graphicx}        
\usepackage{multicol}        

\usepackage{csquotes}        
\usepackage{epsfig}          

\usepackage{caption}         
\usepackage{tabu}            

\usepackage{relsize}

\usepackage{verbatim}

\usepackage{booktabs}

\usepackage{bm}
\usepackage{epstopdf}

\usepackage{mathtools}

\usepackage{chapterbib}
\usepackage{cite}

\usepackage[plainpages=false,       
            hypertexnames=false,    
            bookmarks=false,        
            breaklinks=true,        
            colorlinks=true,      
            citecolor=blue,linkcolor=blue,urlcolor=black]{hyperref}

\makeatletter
\providecommand*{\toclevel@titlech}{0}
\providecommand*{\toclevel@authorch}{0}
\makeatother

\usepackage{amsmath}
\usepackage{amsfonts}
\usepackage{amssymb}

\usepackage{float}

\usepackage[caption=false,farskip=0pt,justification=centerlast,font=scriptsize,labelformat=simple]{subfig}

\usepackage[capitalise]{cleveref}
\usepackage{xspace}
\usepackage[table]{xcolor}
\usepackage{engord}
\usepackage{hhline}
\usepackage{multirow}
\usepackage{varwidth}
\usepackage{overpic}
\usepackage{url}

\usepackage{tabularx}

\usepackage{algpseudocode}
\usepackage[chapter,ruled]{algorithm}
\makeatletter
\renewcommand\floatc@ruled[2]{\captionstyle\strut{\@fs@cfont #1} #2\par}
\renewcommand\fs@ruled{\def\@fs@cfont{\bfseries}\let\@fs@capt\floatc@ruled
  \def\@fs@pre{\hrule height0.8pt depth0pt \kern3pt}%
  \def\@fs@post{\kern2pt\hrule\relax}%
  \def\@fs@mid{\kern2pt\hrule \kern2pt}%
  \let\@fs@iftopcapt\iftrue}
\makeatother

\makeatletter
\renewcommand\subparagraph{\@startsection{paragraph}{4}{\z@}%
                       {-18\p@}
                       {6\p@}
                       {\normalfont\normalsize\itshape\bfseries
                        \rightskip=\z@ \@plus 8em\pretolerance=10000 }}
\makeatother

\makeindex             

\newcommand{\etal}{et~al.\xspace}
\newcommand{\eg}{e.g.\xspace}
\newcommand{\ie}{i.e.\xspace}

\usepackage[normalem]{ulem}

\newdimen\wOne   \wOne=\textwidth
\newdimen\wTwo   \wTwo=\textwidth   \advance\wTwo   by -1\figcapgap \divide\wTwo by 2
\newdimen\wThree \wThree=\textwidth \advance\wThree by -2\figcapgap \divide\wThree by 3
\newdimen\wFour  \wFour=\textwidth  \advance\wFour  by -3\figcapgap \divide\wFour by 4 
\newdimen\wFive  \wFive=\textwidth  \advance\wFive  by -4\figcapgap \divide\wFive by 5
\newdimen\wSix   \wSix=\textwidth   \advance\wSix   by -5\figcapgap \divide\wSix by 6

\usepackage{adjustbox}
\usepackage{booktabs}
\usepackage{threeparttable}

\usepackage{tikz}
\usetikzlibrary{arrows}
\usetikzlibrary{positioning,calc}
\usetikzlibrary{decorations.pathreplacing}
\usetikzlibrary{decorations.markings}
\usetikzlibrary{fit}
\usetikzlibrary{shapes.callouts}
\usetikzlibrary{shapes.geometric}
\usetikzlibrary{matrix}
\usetikzlibrary{backgrounds}
\usepackage{pgfplots}
\pgfplotsset{compat=1.9}

\IfFileExists{graphicscache.sty}{\usepackage[listing]{graphicscache}}{}

\newcommand{\RGBD}{RGB\nobreakdash-D\xspace}

\begin{document}

\frontmatter

\mainmatter

\newcommand{\noi}{\noindent}

\newcommand{\nclasses}{n_c}         
\newcommand{\loss}{L}               
\newcommand{\one}{\textbf{1}}

\newcommand{\netparams}{\boldsymbol{\theta}}    

\newcommand\ChangeRT[1]{\noalign{\hrule height #1}}

\graphicspath{{Bennamoun/}}

\title{RGB-D image-based Object Detection: from Traditional Methods to Deep Learning Techniques \\\vspace{0.82cm}\small{\textit{RGB-D Image Analysis and Processing}}}
\titlerunning{A Survey on RGB-D image-based Object Detection}

\author{Isaac Ronald Ward \and Hamid Laga \and Mohammed Bennamoun}
\institute{Isaac Ronald Ward \at University of Western Australia, \email{ isaac.ward@uwa.edu.au}
\and Hamid Laga \at Murdoch University, Australia, and The Phenomics and Bioinformatics Research Centre, University of South Australia, \email{H.Laga@murdoch.edu.au} 
\and Mohammed Bennamoun \at University of Western Australia, \email{mohammed.bennamoun@uwa.edu.au}}
%
%
\maketitle

\label{Bennamoun} 

\abstract{Object detection from RGB images is a long-standing problem in image processing and computer vision. It has applications in various domains including robotics, surveillance, human-computer interaction, and medical diagnosis. With the availability of low cost 3D scanners, a large number of \RGBD object detection approaches have been proposed in the past years. This chapter provides a comprehensive survey of the recent developments in this field. We structure the chapter into two parts; the focus of the first part is on techniques that are based on hand-crafted features combined with machine learning algorithms. The focus of the second part is on the more recent work, which is based on deep learning. Deep learning techniques, coupled with the availability of large training datasets, have now revolutionized the field of computer vision, including \RGBD object detection, achieving an unprecedented level of performance. We survey the key contributions, summarize the most commonly used pipelines, discuss their benefits and limitations, and highlight some important directions for future research. }

\index{object detection}
\index{deep learning}

\section{Introduction}
\label{WLBsec:introduction}

Humans are able to efficiently and effortlessly detect objects, estimate their sizes and orientations in the 3D space, and recognize their classes. This capability has long been studied by cognitive scientists. It has, over the past two decades, attracted a lot of interest from the computer vision and machine learning communities mainly because of the wide range of applications that can benefit from it. For instance, robots, autonomous vehicles, and surveillance and security systems rely on accurate detection of 3D objects \index{object detection} to enable efficient object recognition\index{object recognition}, grasping\index{object grasping}, manipulation\index{object manipulation}, obstacle avoidance\index{obstacle avoidance}, scene understanding\index{scene understanding}, and accurate navigation\index{scene understanding}. 

Traditionally, object detection\index{object detection} algorithms operate on images captured with RGB cameras\index{camera!rgb}. However, in the recent years, we have seen the emergence of low-cost 3D sensors, hereinafter referred to as \emph{\RGBD sensors}\index{sensor!\RGBD}\index{camera!\RGBD}, that are able to capture depth\index{depth} information in addition to RGB images. Consequently, numerous approaches for object detection from \RGBD images have been proposed. Some of these methods have been specifically designed to detect specific types of objects, \eg humans, faces and cars. Others are more generic and aim to detect objects that may belong to one of many different classes.  This chapter, which focuses on generic object\index{object!generic object} detection from \RGBD images,  provides a comprehensive survey of the recent developments in this field. We will first review  the traditional methods, which are mainly based on hand-crafted features\index{feature!hand-crafted} combined with machine learning\index{machine learning} techniques. In the second part of the chapter, we will focus on the  more recent developments, which are mainly based on deep learning\index{deep learning}.

The chapter is organized as follows; Section~\ref{sec:general} formalizes the object detection problem, discusses the main challenges, and outlines a taxonomy of the different types of algorithms.  Section~\ref{sec:traditional_methods} reviews the traditional methods, which are based on hand-crafted features and traditional machine learning techniques.  Section~\ref{sec:dlm} focuses on approaches that use deep learning techniques. Section~\ref{sec:discussion_dl} discusses some \RGBD-based object detection pipelines and compares their performances on publicly available datasets, using well-defined performance evaluation metrics. Finally, Section~\ref{sec:summary} summarizes the main findings of this survey and discusses some potential challenges for future research.


\section{Problem statement, challenges, and taxonomy}
\label{sec:general}

Object detection from \RGBD images can be formulated as follows; given an \RGBD image, we seek to find  the  location, size, and orientation of objects of interest, \eg cars, humans, and  chairs. The position and orientation of an object is collectively referred to as the \emph{pose}, where the orientation is provided in the form of Euler angles, quaternion coefficients or some similar encoding. The location can be in the form of a 3D bounding box around the visible and/or non-visible parts of each instance of the objects of interest. It can also be an accurate 2D/3D segmentation\index{segmentation!3D segmentation}, \ie the complete shape and orientation even if only part of the instance is visible.   In general, we are more interested in detecting the whole objects, even if parts of them are not visible due to clutter\index{clutter}, self-occlusions\index{self-occlusion}, and occlusion\index{occlusion} with other objects. This is referred to as \emph{amodal} object detection\index{object detection!amodal object detection}. In this section, we discuss the most important challenges in  this field (Section~\ref{sec:challenges}) and then lay down a taxonomy for the state-of-the-art (Section~\ref{sec:taxonomy}). 

\subsection{Challenges}
\label{sec:challenges}

\setlength{\tabcolsep}{8pt} 
\begin{figure}[t]
\centering{
 \begin{tabular}{p{3.4cm} p{3.4cm} p{3.4cm}}
    \includegraphics[width=3.4cm]{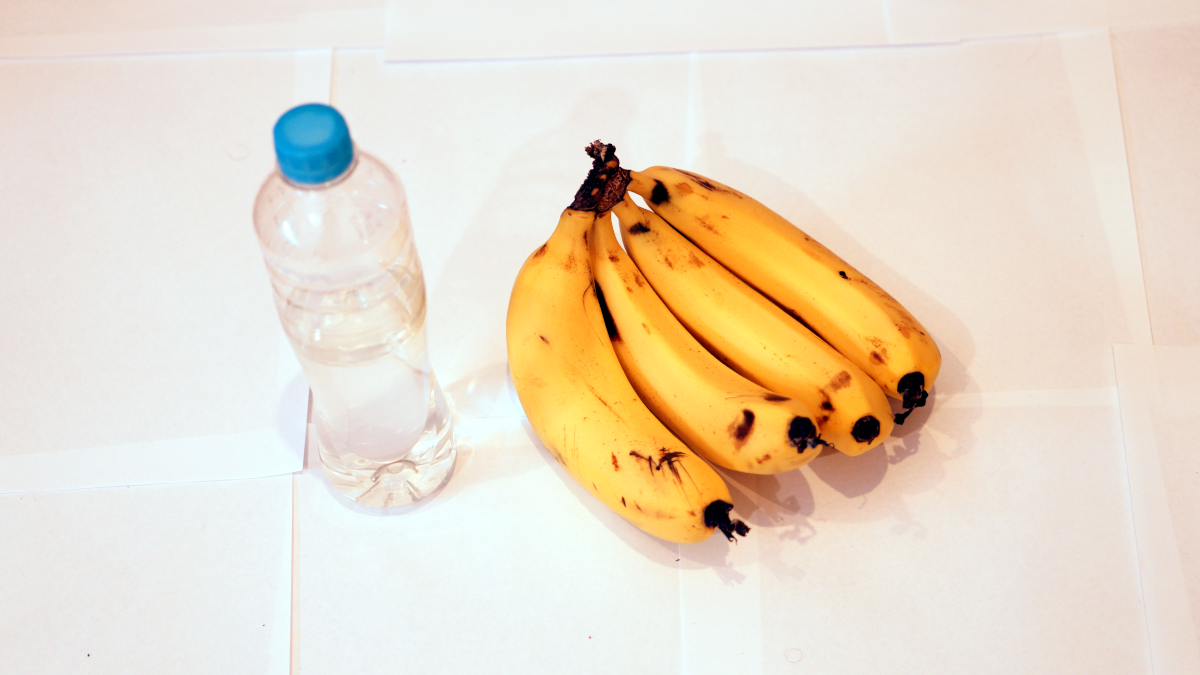} & 
    \includegraphics[width=3.4cm]{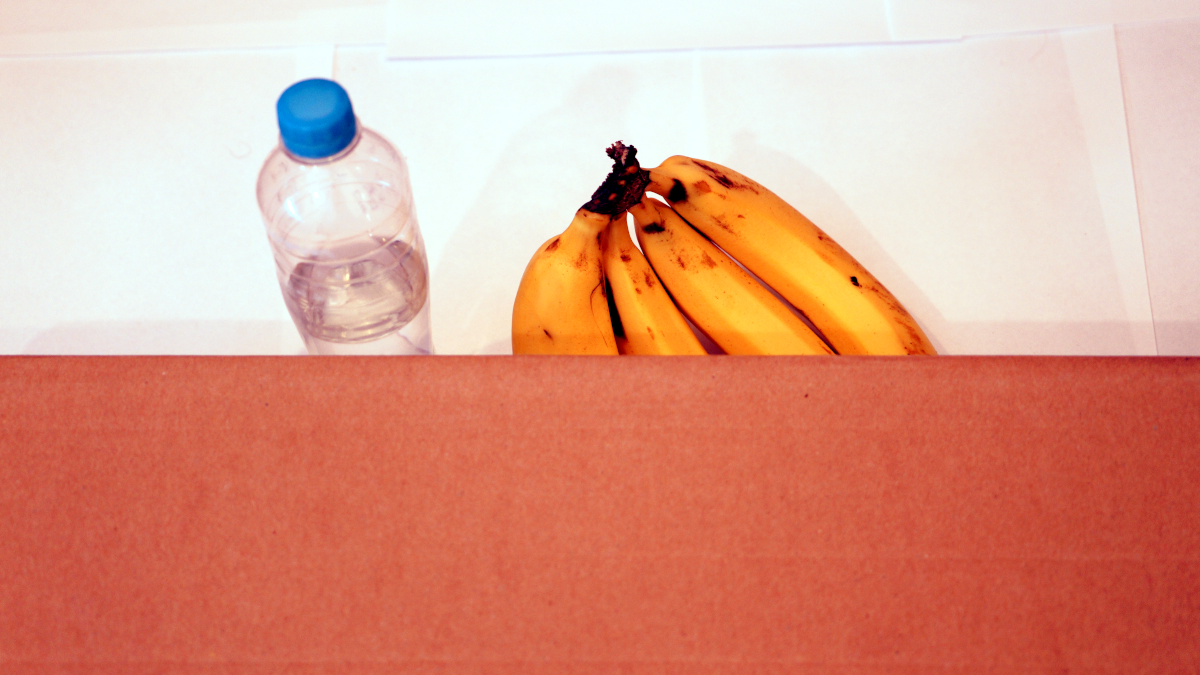} &
    \includegraphics[width=3.4cm]{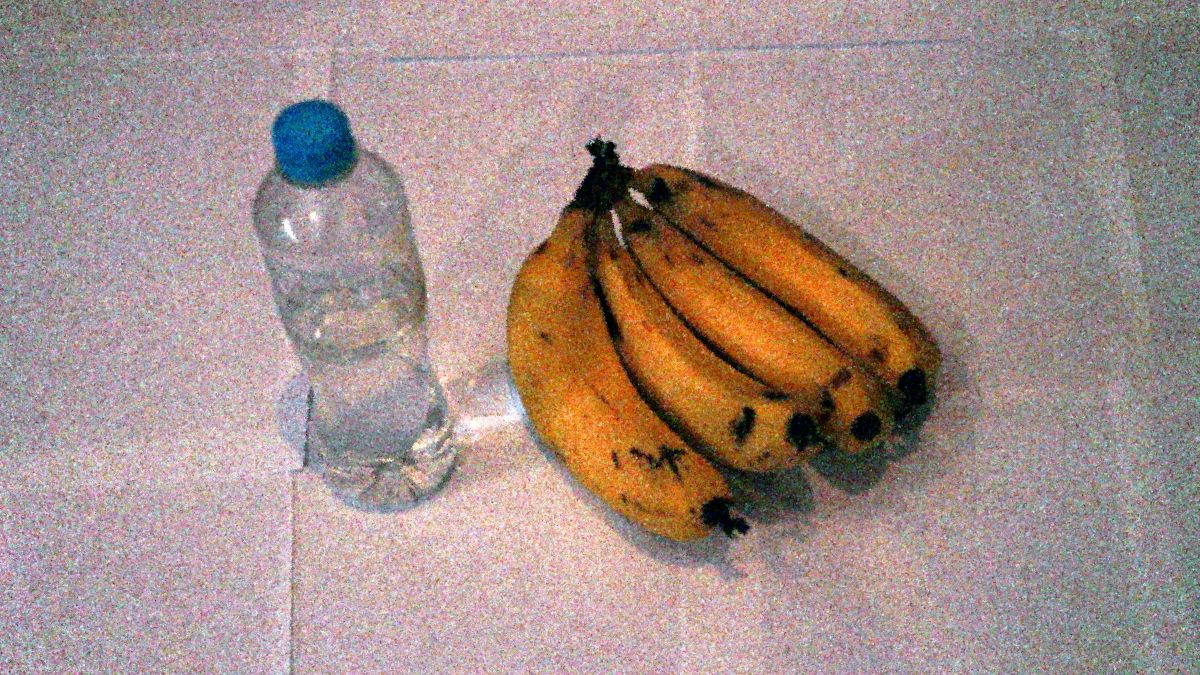} \\
    
    (a)  & 
    (b)  & 
    (c)  \\
    \includegraphics[width=3.4cm]{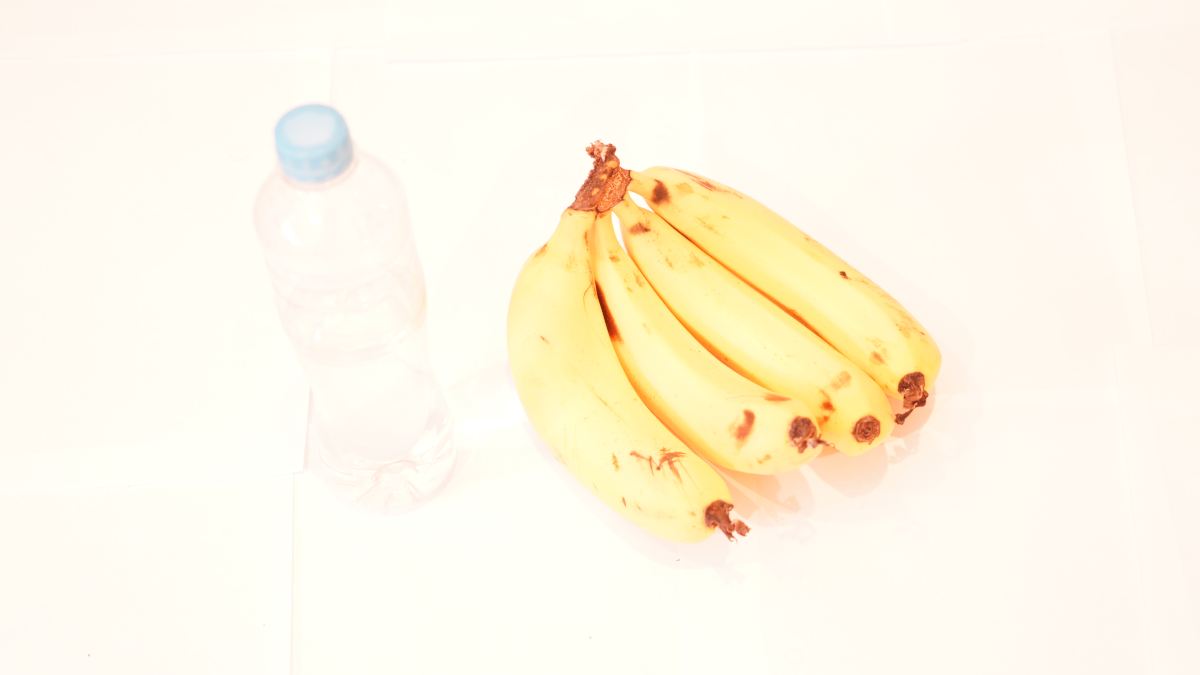} &
    \includegraphics[width=3.4cm]{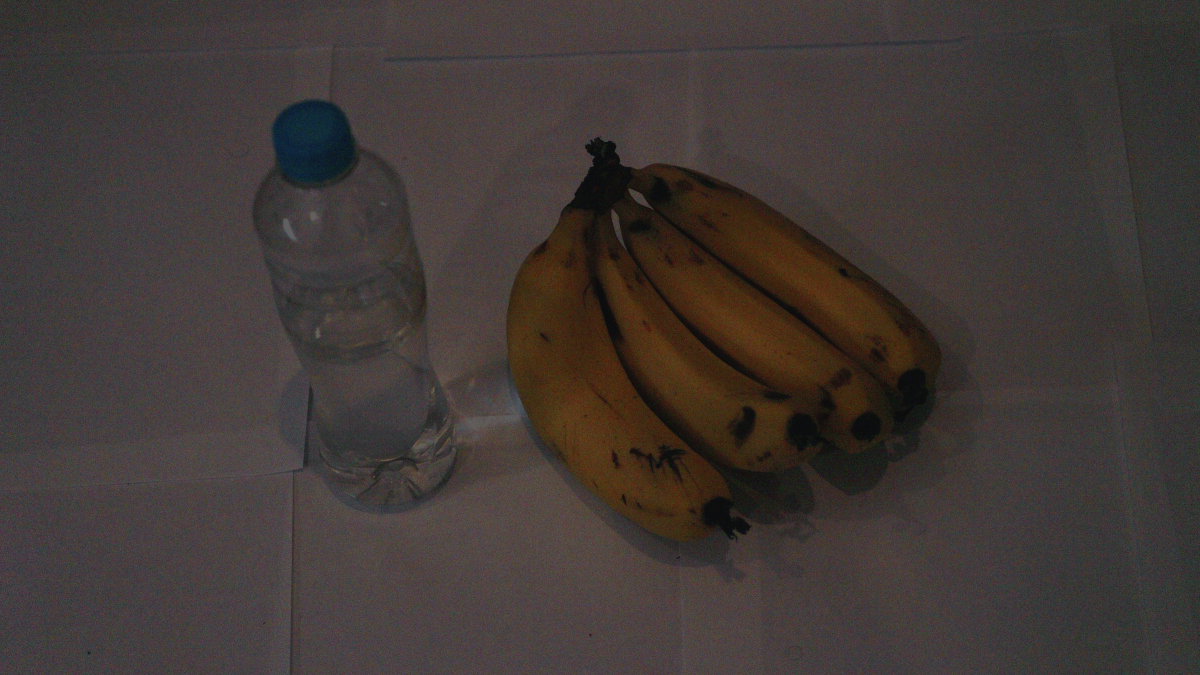} &
    \includegraphics[width=3.4cm]{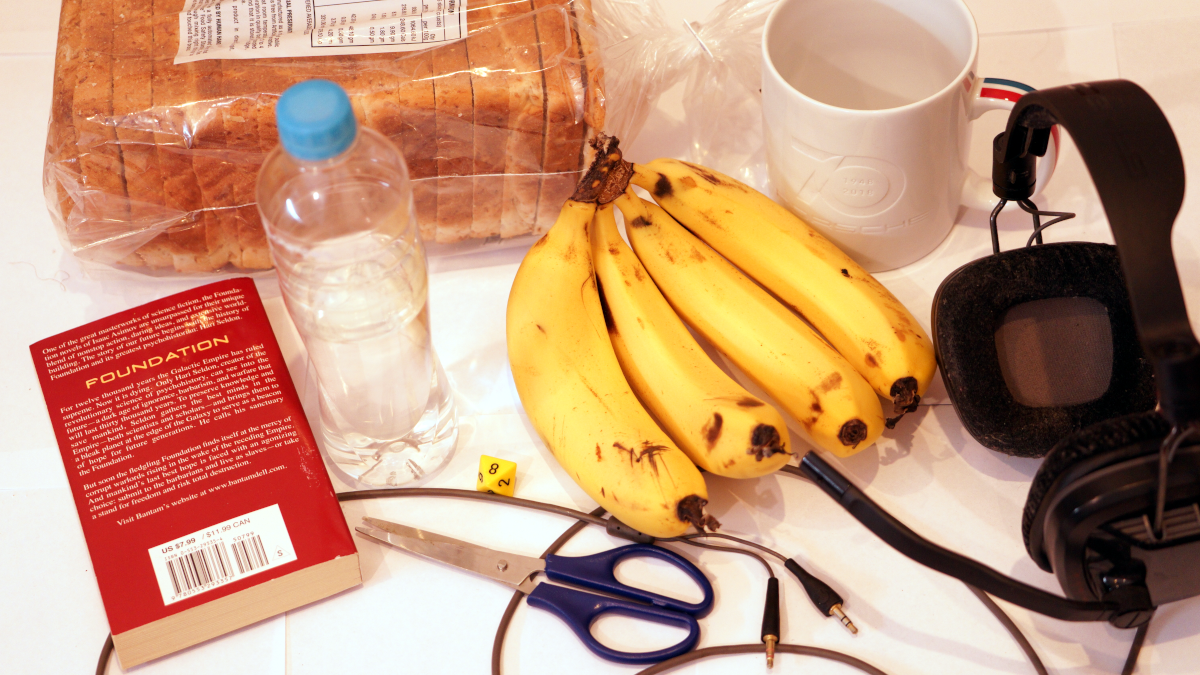} \\ 
    (d)  & 
    (e)  & 
    (f) \\
 \end{tabular}
 \caption{\label{fig:challenges} Illustration of some extrinsic challenges in object detection. (a) Objects of interest are clearly separated from each other and from the uniform background. (b) Two objects partially occluded by a cardboard box. (c) Sensor noise that might affect images. (d)  An overexposed image. (e) An underexposed image.  (f) A cluttered image, which hinders the detection of smaller and occluded objects.}
 }
\end{figure}
\setlength{\tabcolsep}{1.8pt} 

\index{occlusion}
\index{clutter}

Though \RGBD object detection has been extensively investigated, there are a number of challenges that efficient  solutions should address.  Below, we classify these challenges into whether they are due to intrinsic or extrinsic factors\index{intrinsic factors}\index{extrinsic factors}. 
Extrinsic factors refer to all the \textit{external} factors that might affect object detection (see Fig.~\ref{fig:challenges}). Extrinsic challenges include:
\begin{itemize}
    \item \textbf{Occlusions and background clutter. }\index{occlusion}\index{clutter}  The task of object detection algorithms is to not only localize objects in the 3D world, but also to estimate their physical sizes and poses, even if only parts of them are visible in the \RGBD image. In real-life situations, such occlusions can occur at anytime, especially when dealing with dynamic scenes\index{scene!dynamic}.  Clutter can occur in the case of indoor and outdoor scenes\index{scene!indoor}\index{scene!putdoor}.  While biological vision systems excel at detecting objects under such challenging situations,  occlusions and background clutter can significantly affect object detection algorithms.
    
    \item \textbf{Incomplete and sparse data.} Data generated by \RGBD sensors can be incomplete and even sparse in some regions, especially along the $z-$, \ie depth, direction. Efficient algorithms should be able to detect the full extent of the object(s) of interest even when significant parts of it are missing. 
    
    \item \textbf{Illumination. }\index{illumination} \RGBD object detection pipelines should be robust to changes in lighting conditions\index{lighting conditions}.  In fact, significant variations in lighting can  be encountered in indoor and outdoor environments. For  example, autonomously driving\index{autonomous driving} drones\index{drone} and domestic indoor robots\index{robot!indoor robot} are required to operate over a full day-night cycle and are likely to encounter extremes in environmental illumination. As such, the appearance of objects can be significantly affected not only in the RGB image but also in the depth map, depending on the type of 3D sensors used for the acquisition. 
    
    \item \textbf{Sensor limitations. } Though sensor limitations classically refer to colour image noise that occurs on imaging sensors,  \RGBD images are also prone to other unique sensor limitations. Examples include spatial and depth resolution\index{resolution!depth resolution}\index{resolution!spatial resolution}. The latter limits the size of the objects that can be detected.  Depth sensor range limitations are particularly noticeable, \eg the Microsoft Kinect\index{Microsoft Kinect}, which is only sufficiently accurate to a range of approximately $4.5$m \cite{song2015sun}. This prevents the sensor from adequately providing \RGBD inputs in outdoor contexts where more expensive devices, \eg laser scanners\index{laser scanner}, may have to be used~\cite{menze2015kitti}. 
    
    \item \textbf{Computation time. } Many applications, \eg autonomous driving, require real-time object detection.  Despite hardware acceleration\index{hardware acceleration}, using GPUs\index{GPU}, \RGBD-based detection algorithms  can be  slower when compared to their 2D counterparts. In fact, adding an extra spatial dimension increases, relatively, the size of the data. As such, techniques such as sliding windows\index{sliding window} and convolution operations\index{convolution}, which are very efficient on RGB images, become significantly more expensive in terms of computation time and memory storage. 

    \item \textbf{Training data. } Despite the wide-spread use of \RGBD sensors, obtaining  large \textit{labelled} \RGBD datasets to train detection algorithms is more challenging when compared to obtaining purely RGB datasets. This is due to the price and complexity of \RGBD sensors. Although low cost sensors are currently  available, \eg  the Microsoft Kinect, these are usually more efficient in indoor setups. As such, we witnessed a large proliferation of indoor datasets, whereas outdoor datasets are fewer and typically smaller in size.   
\end{itemize}

\setlength{\tabcolsep}{8pt} 
\begin{figure}[t]
\centering{
 \begin{tabular}{p{3.4cm} p{3.4cm} p{3.4cm}}
    \includegraphics[width=3.4cm]{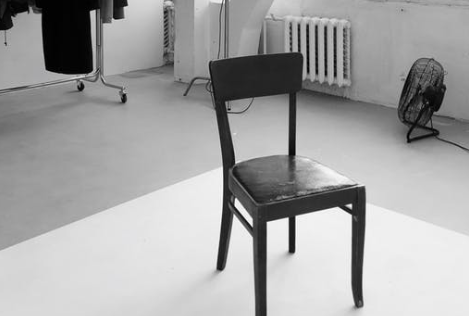} & 
    \includegraphics[width=3.4cm]{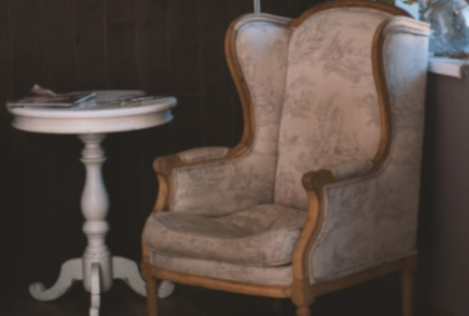} & 
    \includegraphics[width=3.4cm]{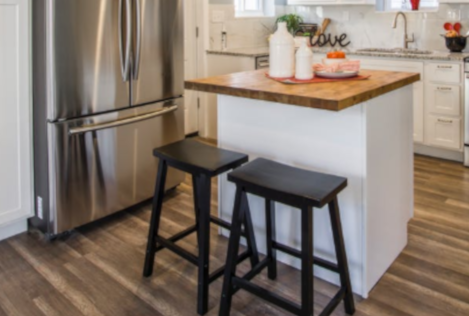} \\
    
    (a)  & 
    (b)  & 
    (c)  \\
    
    \includegraphics[width=3.4cm]{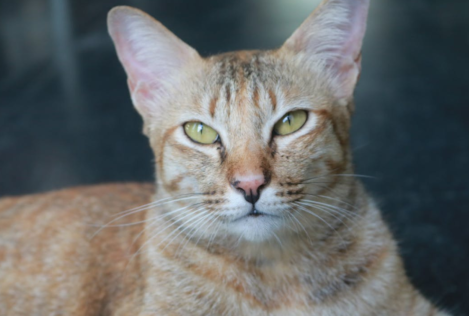} &
    \includegraphics[width=3.4cm]{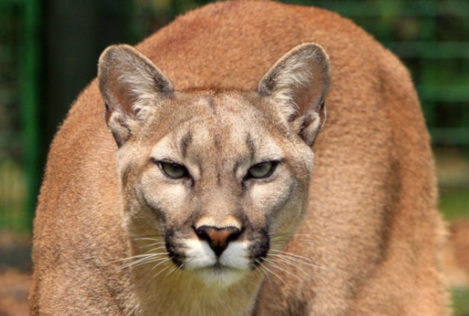} & 
    \includegraphics[width=3.4cm]{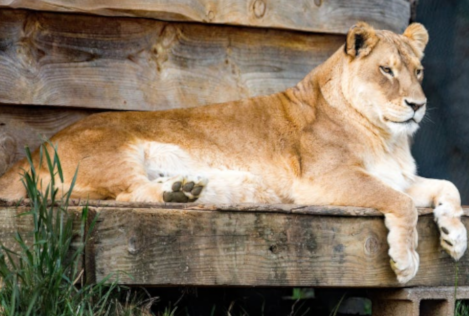} \\
    
    (d)  & 
    (e)  & 
    (f)  \\
    
 \end{tabular}
 \caption{\label{fig:intrchallenges} Illustration of some intrinsic challenges in object detection. (a-c) Intra-class variations where objects of the same class (chair) appear significantly different. (d-f) Inter-class similarities where objects belonging to different classes (cat, cougar \& lion) appear similar. }
 }
\end{figure}
\setlength{\tabcolsep}{1.8pt} 

\noi Intrinsic factors, on the other hand, refer to factors such as deformations\index{deformation}, intra-class variations, and inter-class similarities, which are properties of the objects themselves  (see Fig.~\ref{fig:intrchallenges}):
\begin{itemize}
    \item \textbf{Deformations. } Objects can deform in a rigid\index{deformation!rigid} and non-rigid \index{deformation!non-rigid}way. As such, detection algorithms should be invariant to such shape-preserving deformations.  
    
    \item \textbf{Intra-class variations and inter-class similarities. } \index{intra-class variation}\index{inter-class similarity} Object detection algorithms are often required to distinguish between many objects belonging to many classes. Such objects, especially when imaged under uncontrolled settings, display large intra-class variations.  Also,  natural and man-made objects from different classes may  have strong similarities.  Such intra-class variations and inter-class similarities can significantly affect the performance of the detection algorithms, especially if the number of  \RGBD images used for training is small.   
\end{itemize}

\noi This chapter discusses how the state-of-the-art algorithms addressed some of these challenges.


\subsection{Taxonomy}
\label{sec:taxonomy}

\begin{figure}[t]
\centering
\includegraphics[width=\textwidth]{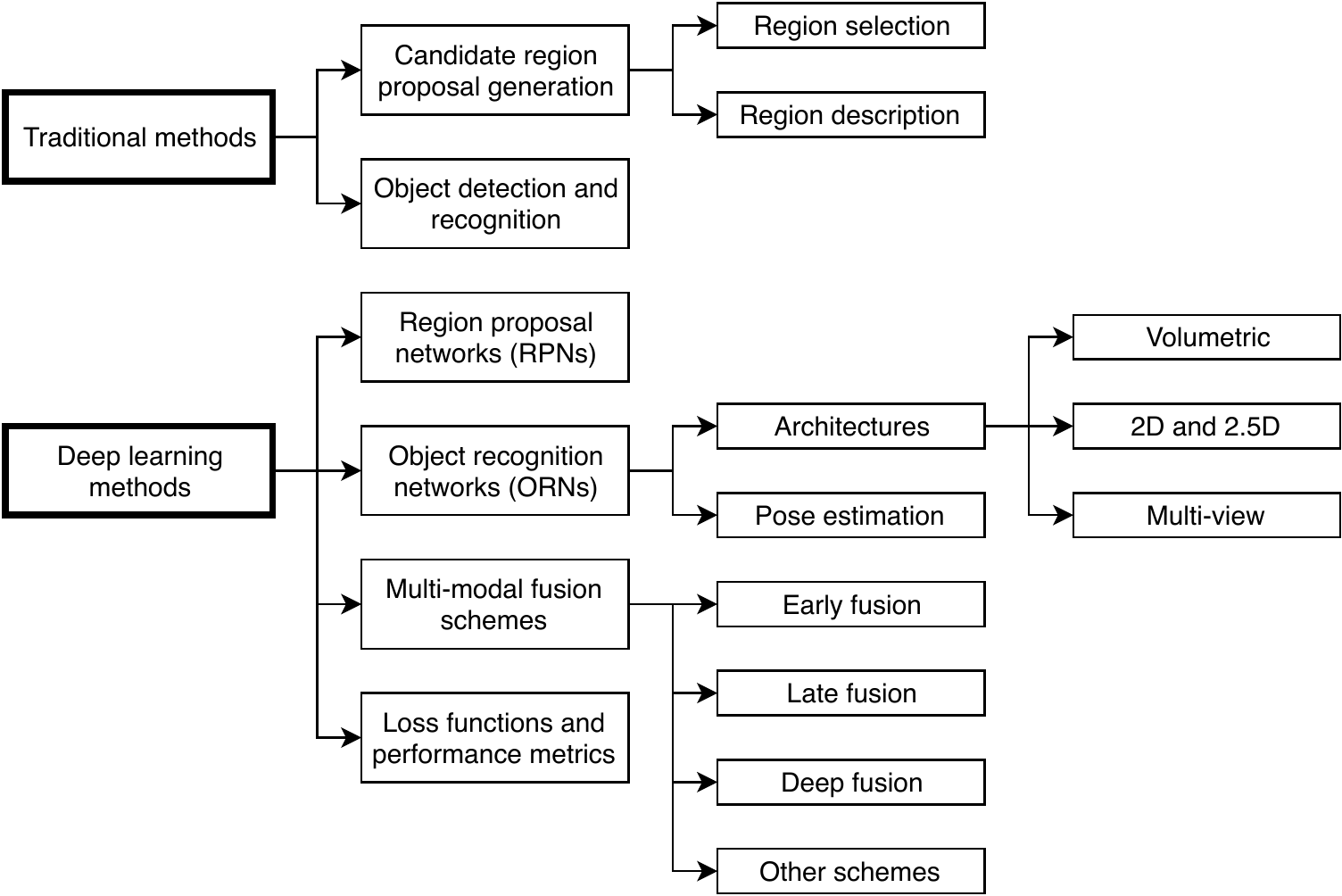}
\caption{Taxonomy of the state-of-the-art traditional and deep learning methods. }
\label{fig:taxonomydiag}
\end{figure}

Figure~\ref{fig:taxonomydiag} illustrates the taxonomy that we will follow for reviewing the state-of-the-art techniques. In particular, both traditional (Section~\ref{sec:traditional_methods}) and deep-learning based (Section~\ref{sec:dlm}) techniques operate in a pipeline of two or three stages. \textbf{The first} stage takes the input \RGBD image(s) and generates a set of region proposals\index{region proposal}. \textbf{The second} stage then refines that selection using some accurate recognition techniques. It also estimates the accurate locations (\ie centers) of the detected objects, their sizes, and their pose. This is referred to as the object's bounding box. This is usually sufficient for applications such as object recognition\index{object recognition} and autonomous navigation\index{autonomous navigation}. Other applications, \eg object grasping \index{object grasping} and manipulation\index{object manipulation}, may require an accurate segmentation of the detected objects. This is usually performed either within the second stage of the pipeline or separately with an additional segmentation\index{segmentation} module, which only takes as input the region within the detected bounding box.

Note that, in most of the state-of-the-art techniques, the different modules of the pipeline operate in an independent manner. For instance, the region proposal module can use traditional techniques based on hand-crafted features\index{feature!hand-crafted}, while the recognition\index{recognition} and localization\index{localization} module can use deep learning\index{deep learning} techniques. 

Another important point in our taxonomy is the way the input is represented and fed into the pipeline. For instance, some methods treat the depth map\index{depth map} as a one-channel image where each pixel encodes depth. The main advantage of this representation is that depth can be processed in the same way as images, \ie using 2D operations\index{2D operations}, and thus there is a significant gain in the computation performance and memory requirements. Other techniques use 3D representations by converting the depth map into either a point cloud \index{point cloud} or a volumetric grid\index{volumetric grid}. These methods, however, require 3D operations \index{3D operation} and thus can be significantly expensive compared to their 2D counterparts.

\textbf{The last} important point in our taxonomy is the fusion \index{fusion} scheme used to merge multi-modal information\index{multi-modal}. In fact, the RGB and D channels of an \RGBD image carry overlapping as well as complementary information. The RGB image mainly provides information about the colour\index{colour} and texture\index{texture} of objects. The depth map\index{depth map}, on the other hand, carries information about the geometry \index{geometry}(\eg size, shape) of objects, although some of this information can also be inferred from the RGB image. Existing state-of-the-art techniques combine this complementary information at different stages of the pipeline (see Fig~\ref{fig:taxonomydiag}).

\section{Traditional Methods}
\label{sec:traditional_methods}
The first generation of algorithms that aim to detect the location and pose of objects in \RGBD images relies on hand-crafted features\footnote{Here we defined hand-crafted or hand-\textit{engineered} features as those which have been calculated over input images using operations which have been explicitly defined by a human designer (\ie \textit{hand-crafted}), as a opposed to learned features which are extracted through optimization procedures in learning pipelines. } \index{feature!hand-crafted} combined with machine learning\index{machine learning} techniques. They operate in two steps: (1) candidate region selection, and (2) detection refinement.



  


\subsection{Candidate region proposals generation}
\label{ssec:candidateselection}
The first step of an object detection algorithm is to generate a set of candidate regions, also referred to as hypotheses, from image and depth cues. The set will form the potential object candidates and should cover the majority of the true object locations.  This step can be seen as a coarse recognition step in which regions are roughly classified into whether they contain the objects of interest or not. While the classification is not required to be accurate, it should achieve a high recall so that object regions will not be missed.


\subsubsection{Region selection}

The initial set of candidate regions can be generated by using (1) bottom-up clustering-and-fitting\index{clustering and fitting}, (2) sliding windows\index{sliding window}, or (3) segmentation \index{segmentation} algorithms.

\textbf{Bottom-up clustering-and-fitting methods} start at the pixel and point level, and iteratively cluster such data into basic geometric primitives \index{geometric primitives} such as cuboids. These primitives can be further grouped together to form a set of complex object proposals\index{object proposal}. In a second stage, an optimal subset is selected using some geometric, structural, and/or semantic cues.  Jiang and Xiao~\cite{jiang2013linear}  constructed a set of cuboid candidates using pairs of superpixels in an \RGBD image.  Their method starts by partitioning the depth map, using both colour and surface normals\index{normal}, into superpixels \index{superpixe} forming piecewise planar patches. Optimal cuboids are then fit to pairs of adjacent planar patches. Cuboids with high fitting accuracy are then considered as potential candidate hypotheses. Representing objects using cuboids has also been used in~\cite{bleyer2012extracting,lin2013holistic,khan2015separating}.  Khan \etal~\cite{khan2015separating} used a similar approach to fit cuboids to pairs of adjacent planar patches and also to individual planar patches. The approach also distinguishes between scene bounding cuboids and object cuboids.   

These methods tend, in general, to represent a scene with many small components, especially if it contains objects with complex shapes. To overcome this issue, Jiang~\cite{jiang2014finding} used approximate convex\index{convex} shapes, arguing that they are better than cuboids in  approximating generic objects\index{object!generic}.

Another approach is to use a \textbf{sliding window method}\index{sliding window} \cite{song2014sliding,Song_2016_CVPR,Deng_2017_CVPR} where detection is performed by sliding, or \textit{convolving}, a window over the image.  At each window location, the region of the image contained in the window is extracted, and these regions are then classified as potential candidates or not depending on the similarity\index{similarity} of that region to the objects of interest. Nakahara \etal~\cite{nakahara2017multiscalewindow} extended this process by using multi-scale\index{multi-scale} windows to make the method robust to variations in the size of objects that can be detected.   Since the goal of this step is to extract candidate regions, the classifier \index{classifier} is not required to be accurate. It is only required to have a high recall\index{recall} to ensure that the selected candidates cover the majority of the true object locations. Thus, these types of methods are generally fast and often generate a large set of candidates.

Finally, \textbf{segmentation-based region selection methods}~\cite{bo2014learning,asif2017cascforest} extract candidate regions by segmenting the \RGBD image into meaningful objects and then considering segmented regions separately. These methods are usually computationally expensive and may suffer in the presence of occlusions\index{occlusion} and clutter\index{clutter}.

\subsubsection{Region description} 
Once candidate regions have been identified, the next step is to describe these regions with features that characterize their geometry and appearance. These descriptors \index{descriptor} can be used to refine the candidate region selection, either by using some supervised recognition techniques\index{supervised learning}, \eg Support Vector Machines~\cite{song2014sliding}\index{SVM}, Adaboost~\cite{schapire2013adaboost}\index{Adaboost}, and hierarchical cascaded forests~\cite{asif2017cascforest}\index{hierarchical cascaded forest},  or by using  unsupervised procedures\index{unsupervised learning}.

In principle, any type of features which can be computed from the RGB image can be used. Examples include colour statistics, Histogram of Oriented Gradients (HOG) descriptor~\cite{dalal2005hogdesc}\index{Histogram of Oriented Gradients}\index{HOG}, Scale-Invariant Feature Transform (SIFT)~\cite{lowe2004sift}\index{Scale-Invariant Feature Transform}\index{SIFT}, the Chamfer distance~\cite{barrow1977chamfer}\index{Chamfer}, and  Local Binary Patterns (LBPs)~\cite{he1990lbp}\index{Local Binary Patterns}\index{LBP}. Some of these descriptors can be used to describe the geometry if  computed from the depth map, by treating depth as a grayscale image. Other examples of 3D features include: 
\begin{itemize}
    \item \textbf{3D normal features. } \index{normal} These are used to describe the orientation of an object's surface. To compute 3D normals, one can pick $n$ nearest neighbors for each point, and estimate the surface normal at that point using principal component analysis (PCA)\index{principal component analysis}\index{PCA}. This is equivalent to fitting a plane and choosing the  normal vector \index{normal}to the surface to be the normal vector to that plane, see~\cite{laga20183d}.

    \item \textbf{Point density features~\cite{song2014sliding}. } It is computed by subdividing  each 3D cell into $n\times n \times n$ voxels and building a histogram of the number of points in each voxel. Song \etal~\cite{song2014sliding} also applied  a 3D Gaussian kernel \index{kernel!Gaussian} to assign a weight to each point, canceling the bias of the voxel discretization. After obtaining the histogram inside the cell, Song \etal~\cite{song2014sliding} randomly pick $1000$ pairs of entries and compute the difference within each pair, obtaining what is called the stick feature~\cite{shotton2013efficient}. The stick feature \index{feature!stick feature} is then concatenated with the original count histogram to form the point density feature\index{feature!point density}. This descriptor captures both the first order (point count) and second order (count difference) statistics of the point cloud~\cite{song2014sliding}.

    \item \textbf{Depth statistics.} This can include the first and second order statistics as well as the histogram of depth. 
    
    \item \textbf{Truncated Signed Distance Function (TSDF)~\cite{newcombe2011kf}. } \index{signed distance function}\index{SDF} For a region divided into $n \times n \times n$  voxels, the TSDF value of each voxel is defined as the signed distance between the voxel\index{voxel} center and the nearest object point on the line of sight from the camera. The distance is clipped to be between $-1$ and $1$. The sign  indicates whether the voxel is in front of or behind a surface, with respect to the camera's line of sight. 
    
    
    \item \textbf{Global depth contrast (GP-D) \cite{ren2015globpriors}. } This descriptor measures  the saliency \index{saliency} of a superpixel \index{superpixel} by considering its depth contrast with respect to all other superpixels.  
 
    \item \textbf{Local Background Enclosure (LBE) descriptor~\cite{Feng_2016_CVPR}. } This descriptor, which is also used to detect salient objects\index{salient objects}, is designed based on the observation that salient objects tend to be located  in front of their surrounding regions. Thus, the descriptor can be computed by creating patches,  via superpixel segmentation~\cite{achanta2012slic},  and considering the angular density of the surrounding patches which have a significant depth difference to the center patch (a difference beyond a given threshold).  Feng \etal~\cite{Feng_2016_CVPR} found that LBE features outperform depth-based features such as anisotropic Center-Surround Difference (ACSD)~\cite{ju2014salientanisotropic}, multi-scale depth contrast (LMH-D)~\cite{Peng_2014_ECCV},  and global depth contrast (GP-D)~\cite{ren2015globpriors}, when evaluated on the RGBD1000~\cite{Peng_2014_ECCV} and NJUDS2000~\cite{ju2014salientanisotropic} \RGBD benchmarks.

    \item \textbf{Cloud of Oriented Gradients (COG) descriptor~\cite{ren2016cogs}.} It extends the HOG \index{HOG} descriptor, which was originally designed for 2D images~\cite{buch2009hog3d, getto2015retrieval}, to 3D data. 
    \item \textbf{Histogram of Control Points (HOCP) descriptor~\cite{sahin2016houghforest, sahin2017partextract}.} A volumetric descriptor calculated over patches of point clouds featuring occluded objects. The descriptor is derived from the Implicit B-Splines (IBS) feature. 
\end{itemize}


\noi In general, these hand-crafted features \index{feature!hand-crafted} are computed at the pixel, super-pixel, point, or patch level. They can also be used to characterize an entire region by aggregating the features computed at different locations on the region using histogram and/or Bag-of-Words \index{Bag-of-Words} techniques, see~\cite{laga20183d}.   For instance, Song \etal~\cite{song2014sliding} aggregate the 3D normal features by dividing, uniformly, the half-sphere into $24$ bins  along the azimuth and elevation angles. Each bin encodes the frequency of the normal vectors\index{normal vector} whose orientation falls within that bin.   
Alternatively, one can use a Bag-of-Words learned from training data to represent each patch using a histogram which encodes the frequency of occurrences of each code-word \index{code-word} in the patch~\cite{song2014sliding}.

\subsection{Object detection and recognition}
Given a set of candidate objects, one can train, using the features described in Section \ref{ssec:candidateselection},  a classifier \index{classifier} that takes these candidates and classifies them into either true or false detections.  Different types of classifiers have been used in the literature. Song and Xiao~\cite{song2014sliding} use Support Vector Machines\index{Support Vector Machines}\index{SVM}. Asif \etal~\cite{asif2017cascforest} use hierarchical cascade forests\index{hierarchical cascade forests}. In general, any type of classifiers, \eg AdaBoost\index{AdaBoost},  can be used to complete this task. However,  in many cases, this approach is not sufficient and may lead to excessive false positives and/or false negatives. Instead, given a set of candidates, other approaches select a subset of shapes that best describes the \RGBD data whilst satisfying some geometrical, structural and semantic constraints. This has been formulated as the problem of optimizing an objective function \index{objective function} of the form \cite{jiang2013linear,jiang2014finding}:
\begin{equation}
    E(x) = E_d(x) + E_r(x),
    \label{eq:box_fitting_refinement}
\end{equation}

\noi where $x$ is the indicator vector\index{indicator vector} of the candidate shapes, \ie  $x(i) =1$ means that the $i-$th shape is selected for the final subset.  Khan \etal~\cite{khan2015separating} extended this formulation to classify \RGBD superpixels\index{superpixel} as cluttered\index{clutter} or non-cluttered, in addition to object detection. As such, another variable $y$ is introduced. It is a binary indicator vector whose $i-$th entry indicates whether the $i-$th superpixel is cluttered or not.   This has also been formulated as the problem of optimizing an objective function of the form:
\begin{equation}
    U(x, y) = E(x) + U(y) + U_c(x, y). 
    \label{eq:khan_energy}
\end{equation}

\noi Here, $E(x)$ is given by Equation~\eqref{eq:box_fitting_refinement}. $U(y)$ also consists of two potentials:
\begin{equation}
    U(y) = U_d(y) + U_r(y),
    \label{eq:energy_2}
\end{equation}
where the unary term $U_d$ is the data likelihood of a superpixel's label, and the pairwise potential $U_r$ encodes the spatial smoothness between neighboring superpixels.  The third term of Equation~\eqref{eq:khan_energy} encodes compatibility constraints, \eg enforcing the consistency of the cuboid labelling and the superpixel labelling. 

\vspace{6pt}
\noi\textbf{The data  likelihood term $E_d$. } This term  measures the cost of matching the candidate shape to the data. For instance, Jiang 
\etal~\cite{jiang2013linear} used cuboids for detection, and defined this term as the cost of matching a cuboid to the candidate shape. On the other hand, Jiang~\cite{jiang2014finding} used convex shapes, and defined $E_d$ as a measure of concavity\index{concavity} of the candidate shape. Several other papers take a learning approach. For instance, Khan \etal~\cite{khan2015separating} computed seven types of cuboid features (volumetric occupancy, colour consistency, normal consistency, tightness feature, support feature, geometric plausibility feature, and cuboid size feature), and then predicted the local matching quality using machine learning approaches.

\vspace{6pt}
\noi\textbf{The data  likelihood term $U_d$. } \index{likelihood}Khan \etal~\cite{khan2015separating} used a unary potential that captures, on each superpixel, the appearance\index{appearance} and texture \index{texture}properties of cluttered and non-cluttered regions. This is done by extracting several cues including image and depth gradient,  colour, surface normals, LBP features, and self-similarity features. Then, a Random Forest classifier was trained to predict the probability of a region being a clutter or a non-clutter.

\vspace{6pt}
\noi\textbf{The regularization terms $E_r$ and $U_r$. }  The second terms of Equations~\eqref{eq:box_fitting_refinement} and~\eqref{eq:energy_2} are regularization terms, which incorporate various types of constraints. Jian~\cite{jiang2014finding} define this term as:
\begin{equation}
    E_r(x) = \alpha N(x) + \beta I(x) - \lambda A(x),
\end{equation}

\noi where  $\alpha, \beta$, and $\lambda$ are weights that set the importance of each term. $N(x)$ is the number of selected candidates, $I(x)$ measures the amount of intersection (or overlap) between two neighboring candidate shapes, and $A(x)$ is the amount of area covered by the candidate shapes projected onto the image plane. Jiang and Xiao~\cite{jiang2013linear}, and later Khan \etal~\cite{khan2015separating}, used the same formulation but added a fourth term, $O(x)$, which penalizes occlusions\index{occlusion}.   

For the superpixel pairwise term of Equation~\eqref{eq:energy_2},  Khan \etal~\cite{khan2015separating} defined a contrast-sensitive Potts model\index{Potts} on spatially neighboring superpixels, which encouraged the smoothness of cluttered and non-cluttered regions. 

\vspace{6pt}
\noi\textbf{The compatibility term $U_c$. }  Khan \etal~\cite{khan2015separating} introduced the compatibility term to link the labelling of the superpixels to the cuboid selection task. This ensures consistency between the lower level and  the higher level of the scene representation. It consists of two terms: a superpixel membership potential, and a superpixel-cuboid occlusion \index{occlusion} potential. The former ensures that a superpixel is associated with at least one cuboid if it is not a cluttered region. The latter ensures that a cuboid should not appear in front of a superpixel which is classified as clutter\index{clutter}, \ie a detected cuboid cannot completely occlude a superpixel on the 2D plane which takes a clutter label.


\vspace{6pt}
\noi\textbf{Optimization. } The final step of the process is to solve the optimization problem of Equations~\eqref{eq:box_fitting_refinement} or~\eqref{eq:energy_2}. Jian~\cite{jiang2014finding} showed that the energy function of Equation~\eqref{eq:box_fitting_refinement} can be linearized and optimized using efficient algorithms such as the branch-and-bound method\index{branch-and-bound}.  Khan \etal~\cite{khan2015separating}, on the other hand, transformed the minimization problem into a Mixed Integer Linear Program (MILP) \index{Mixed Integer Linear Program}\index{MILP} with linear constraints, which can be  solved using the branch-and-bound method.


\section{Deep Learning Methods}
\label{sec:dlm}

Despite the extensive research, the performance of traditional methods is still far from the performance of the human visual system, especially when it comes to detecting objects in challenging situations, \eg highly-cluttered scenes and scenes with high occlusions. In fact, while traditional methods perform well in detecting and producing bounding boxes on visible parts, it is often desirable to capture the full extent of the objects regardless of occlusions \index{occlusion}and clutter\index{clutter}. Deep learning-based  techniques aim to overcome these limitations. They generally operate following the same pipeline as the traditional techniques (see Section~\ref{sec:traditional_methods}), \ie region proposals extraction,  object recognition, and 3D bounding box location and pose estimation. However, they replace some or all of these building blocks with deep learning networks.  This section reviews the different deep learning architectures that have been proposed to solve these problems. Note that these techniques can be combined with traditional techniques; \eg one can use traditional techniques for region proposals\index{region proposal} and deep learning networks for object recognition\index{object recognition}, bounding box location and pose refinement\index{pose}.

\subsection{Region proposal networks}

In this section, we are interested in amodal detection\index{amodal detection}, \ie the inference of the full 3D bounding box beyond the visible parts. This critical step in an object detection pipeline is very challenging since different object categories can have very different object sizes in 3D.  Region Proposal Networks (RPNs)\index{region proposal network} are central to this task since they reduce the search space considered by the remainder of the object detection pipeline.

We classify existing RPNs into three different categories.  \textbf{Methods in the first category} perform the detection on the RGB image and then, using the known camera projection matrix\index{projection matrix},  lift the  2D region  to a 3D frustum that defines a 3D search space for the object.  In general, any 2D object detector, \eg~\cite{long2014fcnseg}, can be used for this task. However, using deep CNNs \index{CNN} allows the extraction of rich features at varying degrees of complexity and dimensionality, which is beneficial for the purpose of \RGBD object detection tasks.  In fact, the success of the recent object detection pipelines  can be largely attributed to the automatic feature learning aspect of convolutional neural networks.  For instance, Qi \etal~\cite{Qi_2018_CVPR} used the Feature Pyramid Networks (FPN)~\cite{lin2017feature}\index{feature pyramid network}, which operate on RGB images, to detect region proposals. Lahoud and Ghanem~\cite{Lahoud_2017_ICCV}, on the other hand,   use the 2D Faster R-CNN~\cite{ren2015fasterrcnn} \index{R-CNN} and VGG-16~\cite{karen2014vgg}\index{VGG}, pre-trained on the  2D ImageNet database \cite{russakovsky2015imagenet}\index{ImageNet},  to position  2D bounding boxes around possible objects with high accuracy and efficiency.  These methods have been applied for indoor and outdoor scenes captured by \RGBD cameras, and for scenes captured using LIDAR sensors~\cite{chen2017multi}\index{LIDAR}.

The rational behind these methods is that the resolution of data produced  by most 3D sensors is still lower than the resolution of RGB images, and that 2D object detectors are mature and quite efficient. RGB-based detection methods, however, do not benefit from the additional information encoded in the depth map. 

\textbf{The second class of methods} aim to address these limitations. They treat depth as an image and perform the detection of region proposals  on the \RGBD image either by using traditional techniques or by using 2D convolutional networks. For instance, Gupta \etal~\cite{gupta2014corr}, and later Deng and Latecki~\cite{Deng_2017_CVPR},  computed an improved contour\index{contour} image from an input \RGBD image. An improved contour image is defined as the contour image but augmented with additional features such as the gradient\index{gradient}, normals\index{normal}, the geocentric pose\index{pose}, and appearance features such as the soft edge map produced by running the contour detector \index{contour detector} on the RGB image.  They then generalize the multiscale combinatorial grouping 
(MCG) algorithm~\cite{arbel2014mcg,pont-tuset2015mcg} to \RGBD images for region proposal and ranking. Note that  both the hand-crafted features\index{feature!hand-crafted} as well as the region  recognition and ranking algorithms can be replaced with deep learning techniques, as in \cite{Qi_2018_CVPR}. 

Chen \etal~\cite{chen2017multi} took the point cloud (produced by LIDAR sensors) and the RGB image, and produced two types of feature maps: the bird's eye view features and the front view features. The bird's eye view representation is encoded by height, intensity, and density. First, the   point cloud\index{point cloud} is projected and discretized into a 2D grid with a fixed resolution. For each cell in the grid, the height feature is computed as the maximum height of the points in that cell. To encode more detailed height information, the point cloud is divided equally into $m$ slices. A height map \index{height map} is computed for each slice, thus obtaining $m$ height maps. The intensity feature is defined as the reflectance\index{reflectance} value of the point which has the highest height in each cell.   Finally, a network that is similar to the region proposal network of~\cite{ren2015faster}  was used to generate region proposals from the bird's eye view map. 

These methods require the fusion of the RGB and depth data. This can be done by  simply concatenating the depth data with the RGB data and using this as input to the RPN. However, depth data encodes geometric information, which is distinct from the spatial and colour information provided by monocular RGB images. As such, Alexandre \etal~\cite{alexandre2014objrec} found that fusing amodal networks\index{amodal networks} with a majority voting\index{voting!majority voting} scheme produced better results in  object recognition tasks, with an improvement of $29\%$ when compared to using simple RGB and depth frame concatenation.  Note that, instead of performing fusion at the very early stage, \eg by concatenating the input modalities, fusion can be performed at a later stage by concatenating features computed from the RGB and D maps, or progressively using the complementarity-aware fusion network  of Chen \etal~\cite{Chen_2018_CVPR}, see Section~\ref{WLBsec:fusion}. 

\textbf{The third class of methods} take a 3D approach. For instance,  Song and Xia~\cite{Song_2016_CVPR} projected both the depth map and the RGB image into the 3D space forming a volumetric scene. The 3D scene is then processed with a fully 3D convolutional network, called a \emph{3D Amodal Region Proposal Network}, which generates region proposals in the form of 3D bounding boxes\index{bounding box} at two different scales. Multi-scale RPNs \index{RPN!multi-scale} allow  the detection of objects of different sizes. It performs  a 3D sliding-window\index{sliding window} search with varying window sizes, and produces an objectness  score for each of the non-empty proposal boxes~\cite{alexe2014whatisanobject}.  Finally,  redundant proposals are removed using non-maximum suppression\index{non-maximum suppression}  with an IoU \index{IoU} threshold of $0.35$ in 3D. Also, the approach ranks the proposals based on their objectness score and only selects the top $2000$ boxes to be used as  input to the object recognition network.  

3D detection can be very expensive since it involves 3D convolutional operations. In fact, it can be more than $30$ times slower than its 2D counterpart. Also, the solution space is very large since it includes three dimensions for the location and two dimensions for the orientation of the bounding boxes.  However, 3D voxel grids\index{voxel grid} produced from depth maps are generally sparse as they only contain information near the shape surfaces. To leverage this sparsity, Engelcke \etal~\cite{engelcke2017vote3deep} extended the approach of Song \etal~\cite{song2014sliding} by replacing the SVM \index{SVM} ensemble with a 3D CNN\index{3D CNN}, which operates on voxelized 3D grids. The key advantage of this approach is that it leverages the sparsity encountered in point clouds to prevent huge computational cost that occurs with 3D CNNs. In this approach, the computational cost  is only proportional  to the number of occupied grid cells rather than to the total number of cells in the discretized 3D grid as in~\cite{wang2015votingfv}.

\subsection{Object recognition  networks}
\index{object recognition network}
\index{ORN}
Once region proposals have been generated, the next step is to classify these regions into whether they correspond to the objects we want to detect or not, and subsequently refine  the detection by estimating the accurate location, extent, and pose\index{pose} (position and orientation) of each object's bounding box. The former is a classification\index{classification} problem, which has been well solved using Object Recognition Networks (ORNs) \cite{schwarz2015pretrain,eitel2015corr,asif2017cascforest}. An ORN takes a candidate region, and assigns to it a class label, which can be binary, \ie $1$ or $0$, to indicate whether it is an object of interest or not, or multi-label where the network recognizes the class of the detected objects. 

There are several ORN architectures that have been proposed in the literature \cite{maturana2015voxnet,Song_2016_CVPR,Deng_2017_CVPR,chen2017multi,Gupta_2015_CVPR}.  In this section, we will discuss some of them based on \textbf{(1)} whether they operate on 2D or 3D (and subsequently whether they are using 2D or 3D convolutions)\index{convolution}, and \textbf{(2)} how the accurate 3D location and size of the bounding boxes are estimated.

%


\subsubsection{Network architectures for object recognition}

\noi\textbf{(1) Volumetric approaches. } The first class of methods are volumetric. The idea is to lift the information in the detected regions into 3D volumes and process them using 3D convolutional networks\index{convolutional neural network}\index{CNN}. For instance, Maturana \etal~\cite{maturana2015voxnet}  used only the depth information to recognize and accurately detect the objects of interest. Their method first converts the point cloud within each 3D region of interest into a $32\times 32 \times 32$  occupancy grid\index{occupancy grid}, with the $z$ axis approximately aligned with gravity. The point cloud is then fed into a 3D convolutional network, termed \emph{VoxNet}\index{VoxNet}, which outputs the class label of the region.

Song and Xia~\cite{Song_2016_CVPR}  followed a similar volumetric approach but they jointly learned the object categories and the 3D box regression from both depth and colour information.  Their approach operates as follows. For each 3D proposal, the 3D volume from depth is fed to a 3D ConvNet\index{ConvNet}, and the 2D colour patch (the 2D projection of the 3D proposal)  is fed to a 2D ConvNet (based on VGG\index{VGG} and pre-trained on ImageNet\index{ImageNet}).  The two latent representations learned by the two networks are then concatenated and further processed with one fully connected layer\index{fully connected layer}.  The network then splits into two branches, each composed of one fully connected layer. The first branch is a classification branch as it produces the class label. The second branch estimates the location and size of the 3D amodal  bounding box\index{amodal bounding box} of the detected object. This approach has two important features; \textbf{first}, it combines both  colour and geometry (through depth) information to perform recognition and regress the amodal bounding box. These two types of information are complementary and thus combining them can improve performance. The \textbf{second} important feature is that it does not directly estimate  the location and size of the bounding box but instead it estimates the residual\index{residual}. That is, it first takes an initial estimate of the size of the bounding box (using some prior knowledge about the class of shapes of interest). The network is then trained to learn the correction that one needs to apply to the initial estimate in order to obtain an accurate location and size of the bounding box.  

\vspace{6pt}
\noi\textbf{(2) 2D and 2.5D approaches. } In the context of object detection, 2D approaches operate over the two spatial dimensions in the input (\ie an RGB image), without exploiting the data encoded in depth images. 2.5D inputs refer to inputs with attached depth images, but importantly, these depth images are treated similarly to how colour images are (without exploiting 3D spatial relationships, \ie using the depth frames as 2D maps where each pixel encodes the depth value). Finally, 3D approaches use the rich spatial data encoded in volumetric or point cloud representations of data (or any other representation which represents the data over three spatial dimensions). 

These 2D and 2.5D approaches are mainly motivated by the performance of the human visual system in detecting and recognizing objects just from partial 2D information. In fact, when the majority of an object area on the depth map is not visible, the depth map only carries partial information.  However, information encoded in the 2D image is rich, and humans can still perceive the objects and estimate their 3D locations and sizes from such images~\cite{Deng_2017_CVPR}. 2D and 2.5D approaches try to mimic the human perception and leverage the 2.5D image features directly using current deep learning techniques. 

In particular, Deng and Latecki~\cite{Deng_2017_CVPR} followed the same approach as Song and Xia~\cite{Song_2016_CVPR}   but  operate on 2D maps using 2D convolutional filters\index{convolutional filter}. The main idea is to regress the 3D bounding box just from the RGB and depth map of the detected 2D regions of interests. Their approach replaces the 3D ConvNet \index{ConvNet} of Song and Xia~\cite{Song_2016_CVPR}  with a 2D ConvNet that processes the depth map. Thus,  it is computationally more efficient than the approaches which operate on 3D volumes, \eg\cite{Deng_2017_CVPR}.

\vspace{6pt}
\noi\textbf{(3) Multi-view approaches. }\index{multi-view} The 2D and 2.5D approaches described above can be extended to operate on multi-view inputs. In fact, many practical systems, \eg autonomous driving, acquire \RGBD data from multiple view points. Central to multi-view techniques is the fusion\index{fusion} mechanism used to aggregate information from different views (see also Section~\ref{WLBsec:fusion}), which can be multiple images and/or depth maps captured from multiple view points.  Some of the challenges which need to be addressed include catering for images gathered at varying resolutions. 

Chen \etal~\cite{chen2017multi} proposed a Multi-View 3D network (MV3D), a region-based fusion network, which combines features from multiple views. The network jointly classifies region proposals and regresses 3D bounding box orientations. The pipeline operates in two stages: multi-view ROI pooling\index{pooling}, and deep fusion\index{deep fusion}. The former is used to obtain feature vectors \index{feature vector} of the same length, since features from different views/modalities usually have different resolutions. The deep fusion network\index{deep fusion network}  fuses multi-view features hierarchically to enable more interactions among features of the intermediate layers from different views.

\subsubsection{Pose estimation}\index{pose}
One of the main challenges in amodal object detection \index{amodal object detection} from \RGBD images is how to accurately estimate the pose of the bounding box of the detected object, even if parts of the objects are occluded.  Early works such as Song and Xia~\cite{Song_2016_CVPR}  do not estimate orientation but  use the major directions of the room in order to orient all proposals. This simple heuristic works fine for indoor\index{indoor scene} scenes, \eg rooms. However, it cannot be easily extended to outdoor scenes \index{outdoor scene} or scenes where no prior knowledge of their structure is known. 

Another approach is to perform an exhaustive search of the best orientations over the discretized space of all possible orientations.  For example,  Maturana \etal~\cite{maturana2015voxnet} performed an exhaustive search over $n=12$ orientations around the $z$ axis and selected the one with the largest activation. At training time, Maturana \etal~\cite{maturana2015voxnet}  augmented the dataset  by creating $n=12$ to $n=18$ copies of each input instance, each rotated $360^o/n$  intervals around the $z$ axis (assuming that the $z$ axis is known). At testing time,  the activations of the output layer over all the $n$ copies are aggregated by pooling\index{pooling}.   This approach, which can be seen as a voting \index{voting} scheme,  has been also used to detect landing zones from LIDAR\index{LIDAR} data~\cite{maturana2015cnnarch}. 

Gupta \etal~\cite{Gupta_2015_CVPR}  considered the problem of fitting a complete 3D object model to the detected objects, instead of just estimating the location, orientation, and size of the object's bounding box. They first detect and segment object instances in the scene and then use a convolutional neural network (CNN)\index{CNN} to predict the coarse pose of the object. Gupta \etal~\cite{Gupta_2015_CVPR} then use the detected region (segmentation mask) to create a 3D representation of the object by projecting points from the depth map. The Iterative Closest Point\index{Iterative Closest Point}\index{ICP} (ICP) algorithm \cite{rusinkiewicz2001icp} is then used to align 3D CAD models to these 3D points.

Finally, some recent approaches regress pose in the same way as they perform recognition, \ie using CNNs. This is usually achieved using a region recognition network\index{region recognition network}, which has two branches of fully connected layers; one for recognition and another one for bounding box regression~\cite{Deng_2017_CVPR,chen2017multi}.  Existing methods differ in the way the bounding boxes are parameterized. For instance, Cheng \etal~\cite{chen2017multi}  represent a bounding box using its eight corners. This is a redundant representation as a cuboid can be described with less information.  Deng and Latecki~\cite{Deng_2017_CVPR} used a seven-entry vector $[x_{cam}, y_{cam}, z_{cam}, l, w, h, \theta]$ where $[x_{cam}, y_{cam}, z_{cam}]$ corresponds to the coordinates of the bounding box's centroid under the camera coordinate system. $[l, w, h]$ represents its 3D size, and $\theta$ is the angle between the principal axis and its orientation vector under the tilt coordinate system. Note  that these methods do not directly regress the pose of the bounding box. Instead,  starting from an initial estimate provided by the Region Proposal Network\index{Region Proposal Network}, the regression network estimates the offset vector, which is then applied to the initial estimate to obtain the final pose\index{pose} of the bounding box.

\subsection{Fusion schemes}
\label{WLBsec:fusion}
\index{fusion}
In the context of \RGBD object detection, we aim to exploit the multiple modalities that are present in  \RGBD images, which carry complementary information (colour and depth data). This, however, requires efficient fusion mechanisms. In this section, we discuss some of the strategies that have been used in the literature. 

\vspace{6pt}
\noi\textbf{(1) Early fusion.} \index{fusion!early fusion}In this scheme, the RGB image and the depth map are concatenated to form a four-channel image~\cite{han2018dlodsurvey}. This happens at the earliest point in the network, \ie before any major computational layers process the image. The concatenated image is then processed with 2D or 3D convolutional filters\index{convolutional filter}. This scheme was adopted in \cite{hou2016shortconns} for the purpose of saliency detection.\index{saliency detection}

\vspace{6pt}
\noi\textbf{(2) Late fusion.} \index{fusion!late fusion} In this scheme, the RGB image and the depth map are processed separately, \eg using two different networks, to produce various types of features. These features are then fused together, either by concatenation or by further processing using convolutional networks.  Eitel \etal~\cite{eitel2015corr}, for example, used two networks, one for depth and one for the RGB image, with each network separately trained on ImageNet~\cite{krizhevsky2012imagenetpretrained}\index{ImageNet}. The feature maps \index{feature map} output by the two networks are then concatenated and presented to a final fusion network, which produces object class predictions. This approach achieved an overall accuracy of  $91.3\% \pm 1.4\%$ on the Washington \RGBD Object Dataset, see Table~\ref{tab:dlperf} for more details regarding pipeline performance. 

Note that, Chen \etal~ \cite{chen2017multi} showed that early and late fusion approaches perform similarly  when tested on the hard category of the KITTI dataset, scoring an average precision \index{average precision} of $87.23\%$ and $86.88\%$.

\vspace{6pt}
\noi\textbf{(3) Deep fusion.} \index{fusion!deep fusion} Early and late fusion schemes are limited in that they only allow the final joint predictions to operate on early or deep representations, so useful information can be discarded. Chen \etal~\cite{chen2017multi} introduced a deep learning fusion scheme, which  fuses features extracted from multiple representations of the input.  The fusion pipeline uses element-wise mean pooling \index{pooling} operations instead of simple concatenations (as in early or late fusion). Chen \etal~\cite{chen2017multi} showed that this fusion mechanism improved performance by about $1\%$ compared to early and late fusion.



\vspace{6pt}
\noi\textbf{(4) Sharable features learning and complementarity-aware fusion. } \index{fusion!complementarity-aware fusion} The fusion methods  described above either learn features from colour and depth modalities separately, or simply treat \RGBD as a four-channel data. Wang \etal~\cite{Wang_2015_ICCV} speculate that  different modalities should contain not only some modal-specific patterns but also some shared common patterns. They then propose a multi-modal feature learning framework for \RGBD object recognition. First, two deep CNN layers are constructed, one for colour and  another for depth. They are then  connected with multimodal layers, which fuse colour and depth information by enforcing a common part to be shared by features of different modalities. This produces features reflecting shared properties as well as modal-specific properties from different modalities. 

Cheng \etal~\cite{Chen_2018_CVPR} proposed a fusion mechanism, termed \emph{complementarity-aware (CA) fusion}, which encourages the determination of complementary information from the different modalities at different abstraction levels.  They introduced a CA-Fuse module, which enables cross-modal\index{cross-modal}, cross-level connections and modal/level-wise supervisions, explicitly encouraging the capture of complementary information from the counterpart, thus reducing fusion ambiguity and increasing fusion efficiency.

\subsection{Loss functions}
\index{loss}

In general, the region proposal network (RPN)\index{region proposal network}\index{RPN} and the object recognition network (ORN) \index{object recognition network}\index{ORN}  operate separately. The RPN first detects candidate regions. The ORN then refines the detection by discarding regions that do not correspond to the objects of interest. The ORN then further refines the location, size, and orientation of the bounding boxes.  As such, most of the state-of-the-art techniques train these networks separately using separate loss functions.

Loss functions inform the network on how poorly it completed its designated task over each training batch using a scalar metric (referred to as the loss, cost, or inverse fitness). The calculation of the loss should incorporate error that the algorithm accumulated during the completion of its task, as the network will change its weights in order to reduce the loss, and thus the error. For example, in object classification networks, the loss might be defined as the mean squared error between the one-hot encoded ground truth labels, and the network's output logits. For object detection networks, the IoU (see Fig.~\ref{fig:iou-prec-rec}) of the detected region and the ground truth region may be incorporated. In this way the loss function design is task-dependant, and performance increases have been observed to be contingent on the loss function's design \cite{ward2019lossfns}. Typically loss functions are hand-crafted, though weightings between terms can be learned \cite{kendall2017geo}. Naturally, numerous loss functions have been devised to train networks to accomplish various tasks \cite{khan2018cnnguide}.

A common loss function that has been used for classification (in the RPN as well as in the ORN) is the softmax\index{softmax} regression loss. Let $\netparams$  be the parameters of the network, $m$ the number of training samples, $\nclasses$ the number of classes,  and $y_i \in \{1, \cdots, \nclasses \}$   the output of the network for the training sample $x_i$. The softmax \index{softmax}regression loss is defined as:
\begin{equation}
    \loss(\netparams) = -\sum_{i=1}^{m}\sum_{k=1}^{\nclasses} \one(y_i = k)\log\left(p(y^i = k | x^i; \netparams)\right).
\end{equation}

\noi Here, $\one(s)$ is equal to $1$ if the statement $s$ is true and $0$ otherwise.  This loss function has been used by Gupta \etal~\cite{Gupta_2015_CVPR} to train their region proposal network, which also provides a coarse estimation of each object's pose. 

Song and Xia~\cite{Song_2016_CVPR}, on the other hand,  trained their multiscale region proposal network \index{region proposal network!multiscale} using a loss function that is a weighted sum of two terms: an objectness term and a box regression term:
\begin{equation}
	\loss(p, p^*, \textbf{t}, \textbf{t}^*) = L_{cls}(p, p^*) + \lambda p  L_{1\_smooth}(\textbf{t}, \textbf{t}^*), 
\end{equation}  

\noi where  $p^*$ is the predicted probability of this region being an object and $p$ is the ground truth, $L_{cls}$ is  the log loss over the two categories (object vs. non-object)~\cite{girshick2015fast}, and $\textbf{t}$ is a 6-element vector, which defines the location and scale of the bounding box of the region. 
The second term, which is the box regression loss term, is defined using the smooth $L_1$ function as follows:
\begin{equation}
	L_{1\_smooth}(x) = \begin{dcases} 
						0.5 x^2 \text{ if } |x| < 1, \\
							 	|x| - 0.5, \text{ otherwise. } 
					\end{dcases}
\end{equation}

\noi Song and Xia~\cite{Song_2016_CVPR} also used a similar loss function to train their ORN. The only difference is in the second term, which is set to zero when the ground-truth probability $p$ is zero. This is because there is no notion of a ground-truth bounding box for background RoIs. Hence, the box regression term is ignored. Finally, Maturana \etal~\cite{maturana2015voxnet} used the multinomial negative log-likelihood\index{multinomial negative log-likelihood} plus $0.001$ times the $L_2$ weight norm for regularization\index{regularization}. 



\section{Discussion and comparison of some pipelines}
\label{sec:discussion_dl}

In this section, we discuss some pipelines for object detection from \RGBD data and compare their performance on  standard benchmarks.   We will first review examples of the datasets that have been used for training and testing the techniques (Section~\ref{sec:datasets}), discuss  different performance evaluation metrics (Section~\ref{sec:metrics}), and finally compare and discuss the performance of some of the key \RGBD-based object detection pipelines (Section~\ref{sec:comparison}).

\subsection{Datasets}
\label{sec:datasets}

Many of the state-of-the-art algorithms rely on  large datasets to train their models and evaluate their performance. Both traditional machine learning and advanced deep learning approaches require labelled datasets in the form of  \RGBD images and their corresponding ground-truth labels. The labels can be in the form of 2D bounding boxes highlighting the object regions in the RGB image and/or the depth map, oriented 3D bounding boxes (3DBBX) delineating the 3D regions of the objects, and/or exact segmentations (in the form of segmentation masks) of the objects of interest.  

\begin{table*}[t]
	\caption{\label{tab:datasets}Examples of datasets used for training and evaluating  object detection pipelines from  \RGBD images.}
    {
		\begin{tabular}{| p{1.6cm} | p{5.0cm} | p{5.0cm}|} 
		\hline 
		\textbf{Name} & \textbf{Description} & \textbf{Size}  \\ 
		\hline 
		KITTI 2015~\cite{menze2015kitti} &
		Cluttered driving scenarios recorded in and around Karlsruhe in Germany. &
		$400$ annotated dynamic scenes from the raw KITTI dataset. 
		\\
		\hline
		KITTI 2012~\cite{geiger2012kitti} &
		As above. &
		$389$ image pairs and more than $200,000$ 3D object annotations. 
		\\
		\hline
		SUN \RGBD~\cite{song2015sun} &
		Indoor houses and universities in North America and Asia. &
		$10,335$ images, $800$ object categories, and $47$ scene categories annotated with $58,657$ bounding boxes (3D).
		\\
		\hline
		NYUDv2 \cite{silberman2012nyu} &
		Diverse indoor scenes taken from three cities. &
		$1,449$ \RGBD images over $464$ scenes. 
		\\
		\hline
		PASCAL3D+ \cite{xiang2014beyond} &
		Vehicular and indoor objects (augments the PASCAL VOC dataset \cite{pascalvocchallenge}). &
		$12$ object categories with $~3,000$ instances per category. 
		\\
		\hline
		ObjectNet3D \cite{xiang2016objectnet3d} &
		Indoor and outdoor scenes. &
		$90,127$ images sorted into $100$ categories. $201,888$ objects in these images and $44,147$ 3D shapes. 
		\\
		\hline
		RGBD1000 \cite{Peng_2014_ECCV} &
		Indoor and outdoor scenes captured with a Microsoft Kinect. &
		$1,000$ \RGBD images. 
		\\
		\hline
		NJUDS2000 \cite{ju2014salientanisotropic} &
		Indoor and outdoor scenes. &
		$2,000$ \RGBD images. 
		\\
		\hline
		LFSD \cite{li2017lightfield} &
		Indoor ($60$) and outdoor ($40$) scenes captured with a Lytro camera. &
		$100$ light fields each composed from raw light field data, a focal slice, an all-focus image and a rough depth map. 
		\\ 
		\hline
		Cornell Grasping \cite{cornellgraspingdataset}  &
		Several images and point clouds of typical graspable indoor objects taken at different poses. &
		$1,035$ images of $280$ different objects. 
		\\
		\hline
		ModelNet10 \cite{wu2015shapenets} &
		Object aligned 3D CAD models for the $10$ most common object categories found in the SUN2012 database \cite{xiao2012sun}. &
		$9,798$ total instances split amongst $10$ object categories, each with their own test/train split.
		\\
		\hline
		ModelNet40 \cite{wu2015shapenets} &
		Object aligned 3D CAD models for $40$ common household objects. &
		$12,311$ total instances split amongst $40$ object categories, each with their own test/train split.
		\\
		\hline
		Caltech-101 \cite{feifei2004caltech101} &
	    Single class object centric images with little or no clutter. Most objects are presented in a stereotypical pose. &
		$9,144$ images sorted into $101$ categories, with $40$ to $800$ images per category. 
		\\
		\hline
		Caltech-256 \cite{griffin2007caltech256} &
	    Single class images with some clutter. &
		$30,607$ images sorted into $256$ categories with an average of $119$ images per category. 
		\\
		\hline
		Washington \RGBD~\cite{lai2013rgb} &
		Turntable video sequences at varying heights of common household objects. &
		$300$ objects organized into $51$ categories. Three video sequences per object.
		\\
        \hline
        \end{tabular}
    }
\end{table*}

Table~\ref{tab:datasets} summarizes the main datasets and benchmarks that are currently available in the literature. Note that several types of 3D sensors \index{sensor!3D} have been used for the acquisition of these datasets. For instance, the SUN \RGBD dataset was constructed using four different sensors: the  Intel RealSense 3D Camera,  the Asus Xtion LIVE PRO, and Microsoft Kinect v1 and v2. Intel Asus and Microsoft Kinect v1 sensors use infrared (IR)\index{infrared} light patterns to generate quantized depth maps, an approach known as \emph{structured light}\index{structure light},  whereas Microsoft Kinect v2 \index{Kinect} uses time-of-flight\index{time-of-flight} ranging. These sensors are suitable for indoor scenes since their depth range is limited to a few meters. On the other hand, The KITTI dataset, which includes outdoor scene categories, has been captured using a Velodyne HDL-64E rotation 3D laser scanner\index{laser scanner}.

Note that some datasets, such as the PASCAL3D+, are particularly suitable for testing the robustness of various algorithms to occlusions\index{occlusion}, since an emphasis was placed on gathering data with occlusions. 

\label{sec:metrics}
\subsection{Performance criteria and metrics}

Object detection usually involves two tasks; the first is to assess whether the object exists in the \RGBD image (classification)\index{classification} and the second is to exactly localize the object in the image (localization)\index{localization}. Various metrics have been proposed to evaluate the performance of these tasks. Below, we discuss the most commonly used ones, see also~\cite{laga20193d}.

\vspace{6pt}
\noi\textbf{Computation time. } Object detection algorithms operate in two phases; the training phase and the testing phase. While, in general, algorithms can afford having a slow training phase, the computation time at runtime is a very important performance criterion. Various applications may have different requirements. For instance, time-critical applications such as autonomous driving\index{autonomous driving} and surveillance\index{surveillance} systems should operate in real time.  Other applications, \eg offline indexing of \RGBD images and videos, can afford slower detection times. However, given the large amount of information they generally need to process, real-time detection is desirable. Note that, there is often a trade-off between computation time at runtime and performance.

\begin{figure}[t]
 \centering
 \begin{tabular}{@{}c@{ }c@{}}
    \includegraphics[height=.25\textwidth]{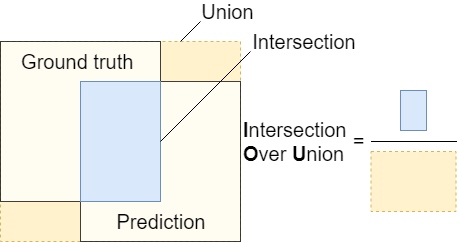} &
    \includegraphics[height=.22\textwidth]{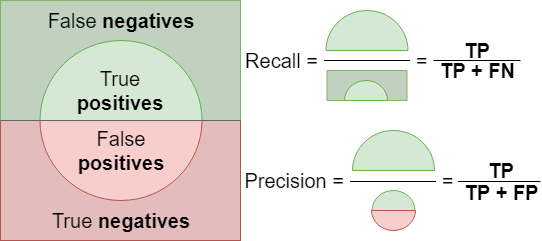} \\
    (a) Intersection over Union (IoU). & (b) Precision-recall.
 \end{tabular}
 \caption{(a) Illustration of the Intersection over Union (IoU)\index{IoU} metric in 2D.    (b) Illustration of how precision and recall are calculated from a model's test results. Recall is the ratio of correct predictions to total objects in the dataset. It measures how complete the predictions are. Precision is the ratio of correct predictions to total predictions made, \ie how correct the predictions are. }
 \label{fig:iou-prec-rec}
\end{figure}

\vspace{6pt}\noi\textbf{Intersection over Union (IoU). }\index{IoU} It measures the overlap between a ground-truth label and a prediction as a proportion of the union of the two regions, see Fig. \ref{fig:iou-prec-rec}. IoU is a useful metric for measuring the predictive power of a 2D/3D  object detector\index{object detector}. IoU thresholds are  applied to sets of detections in order to precisely define  what constitutes a positive detection. For example, an IoU $> 0.5$ might be referred to as a positive detection. Such thresholds are  referred to as \textit{overlap criterion}\index{overlap criteria}.


\vspace{6pt}\noi\textbf{Precision-recall curves. } Precision and recall \index{precision-recall}   are calculated based on a model's test results, see Fig. \ref{fig:iou-prec-rec}.  The precision-recall curve is  generated by varying the threshold which determines what is counted as a positive detection of the class. The model's precision at varying recall values are then plotted to produce the curve. 

\vspace{6pt}\noi\textbf{Average Precision (AP). }\index{average precision} It is defined as  the average value of precision  over the interval from recall $r= 0$ to $r = 1$, which is equivalent to measuring the area under the precision-recall curve ($r$ here is the recall):
\begin{equation}
    AP = \int_0^1 \text{precision}(r)dr.
    \label{eq:average_precision}
\end{equation}

\vspace{6pt}\noi\textbf{Mean Average Precision (mAP) score.  } \index{mean average precision}  It is defined as  the mean average precision over all classes and/or over all IoU \index{IoU} thresholds. 


\vspace{6pt}
\noi\textbf{F- and E-Measures. } \index{F-measure}\index{E-measure} These are two measures which combine  precision and  recall  into a single number to evaluate the retrieval performance\index{retrieval performance}.  The F-measure is the weighted harmonic mean \index{harmonic mean} of precision and recall. It is defined as
\begin{equation}
    F_{\alpha} = \frac{(1+\alpha)\times \text{precision} \times \text{recall}}{\alpha \times \text{precision} + \text{recall}},
    \label{eq:f_measure}
\end{equation}
where $\alpha$ is a weight. When $\alpha=1$ then
\begin{equation}
  F_1 \equiv F = 2\times\frac{\text{precision} \times \text{recall}}{\text{precision} + \text{recall}}.
\end{equation}

\noi The E-Measure\index{E-measure} is defined as $E = 1- F$, which is equivalent to
\begin{equation}
    E = 2 \left(\frac{1}{\text{precision}}+\frac{1}{\text{recall}}\right)^{-1}.
\end{equation}
                    
\noi Note that the maximum value of the E-measure is $1.0$ and the higher the E-measure is, the better is the detection algorithm. The main property of the E-measure is that it quantifies how good are the  results  etrieved in the top of the ranked list. This is very important since, in general, the user of a search engine is more interested in the first page of the query results than in the later pages.

\vspace{6pt}
\noi\textbf{Localization performance. }\index{localization} The localization task is typically evaluated using  the Intersection over Union threshold (IoU)\index{IoU} as discussed above. 

\begin{table*}[t]
	\caption{Performance of some traditional \RGBD-based object detection methods across multiple datasets and performance metrics. All values taken directly from the cited publications. \emph{Class} refers to classification accuracy, \emph{Reco} to recognition accuracy and \emph{mAP} to mean Average Precision. All metrics are measured in percentages except F-score, which is in the range $[0,1]$.}
	\label{tab:tradperf}
	
	\begin{tabular}{@{}| p{2.8cm} | p{1.7cm} | c | c | c | c | c | c | c | c | @{}}
    \ChangeRT{1pt}
    
    \textbf{Method}  & \textbf{Dataset}  &  \multicolumn{2}{c|}{\textbf{Detection rate}} & \textbf{Recall} & & & \textbf{mAP} & \\
                          \cline{3-4} 
            &          &  \textbf{IoU>.7} &  \textbf{IoU>.75} & \textbf{IoU>.5} & \textbf{Class.} &  \textbf{Reco.} & \textbf{IoU>.25} & \textbf{F-score}\\
    \ChangeRT{1pt}
    
    Linear Cuboid Matching \cite{jiang2013linear}           & NYUDv2 & 75.0 & & & & & & \\
    \hline
    Extended CPMC \cite{lin2013holistic}                    & NYUDv2 & & & 42.8 & 60.5 & & & 0.36 \\
    \hline
    Convex Shapes \cite{jiang2014finding}                   & NYUDv2 & & 78.2 & & & & & \\
    \hline
    Sliding Shapes \cite{song2014sliding}                   & NYUDv2$^1$ & & & & & & 39.6 & \\
    \hline
    \textbf{DSS \cite{Song_2016_CVPR}}                      & NYUDv2$^1$ & & & & & & 72.3 & \\
    \hline
    \textbf{DSS \cite{Song_2016_CVPR}}                      & NYUDv2$^2$ & & & & & & 36.3 & \\
    \hline
    Amodal Detection \cite{Deng_2017_CVPR}                  & NYUDv2$^2$ & & & & & & 40.9 & \\
    
    \ChangeRT{1pt}
    
   \textbf{DSS \cite{Song_2016_CVPR}}                       & SUN \RGBD & & & & & & 29.6 & \\
    \hline
    CoG \cite{ren2016cogs}                                  & SUN \RGBD$^3$ & & & & & & 47.6 & \\
    
    \ChangeRT{1pt}
    
    Separating objects / clutter~\cite{khan2015separating}  & RMRC 2013 & 38.0 & & & & & &\\
    \hline
    Sliding Shapes \cite{song2014sliding}                   & RMRC 2013 & & & & & & 62.4 & \\
    
    \ChangeRT{1pt}
    
     M-HMP \cite{bo2014learning}                            & Caltech Bird-200  & & & & & 30.3 & & \\
    \hline
     M-HMP \cite{bo2014learning}                            & Caltech-256 & & & & 58.0 & & &\\
     
    \ChangeRT{1pt}
    
    M-HMP \cite{bo2014learning}                             & MIT Scene-67 & & & & 51.2 & & &\\
    \hline
    STEM-CaRFs \cite{asif2017cascforest}                    & Washington & & & & & 97.6 & & \\
    \hline
    STEM-CaRFs \cite{asif2017cascforest}                    & Cornell grasping & & & & & 94.1 & & \\
    \hline
    GP \cite{ren2015globpriors}                  & NLPR & & & & & & & 0.72 \\
    \hline
    GP \cite{ren2015globpriors}                  & NJUDS400 & & & & & & & 0.76 \\
    \hline
    LBE \cite{Feng_2016_CVPR}                               & NJUDS2000 & & & & & & & 0.72 \\
    \hline
    LBE \cite{Feng_2016_CVPR}                               & RGBD1000 & & & & & & & 0.73 \\
    \hline
    3DHoG \cite{buch2009hog3d}                              & i-LIDS$^4$  & & & & 92.1 & & & \\
    
    \ChangeRT{1pt}
    \end{tabular}
    
    $^1$ Considering only 5 categories from the NYUDv2 dataset.
    $^2$ 3D annotations for the NYUDv2 dataset were improved in \cite{Deng_2017_CVPR} and this improved dataset was used to calculate performance. 
    $^3$ Considering only 10 categories from the SUN \RGBD dataset.
    $^4$ Scenario 1 of the i-LIDS dataset. 
\end{table*}

\begin{table*} 
	\caption{Performance of some deep learning-based object detection methods across multiple datasets and performance metrics. All values taken directly from the original publications. \emph{mAP} refers to mean Average Precision, \emph{Class} refers to classification accuracy, \emph{Valid} to validation set,  and \emph{Seg} to instance segmentation. All metrics are measured in percentages except F-score, which is in the range $[0,1]$.}
	\label{tab:dlperf}
	
	\begin{tabular}{|p{3.2cm} |  p{2.0cm} | c|c|c | c | c | c | c | c |}
    \ChangeRT{1pt}
    \textbf{Method}  & \textbf{Dataset}  &  \multicolumn{3}{c|}{\textbf{Detection mAP}} & & & & \textbf{F-score} \\
                          \cline{3-5}
            &          & \textbf{IoU>.25} & \textbf{IoU>.5} &  \textbf{IoU>.7} & \textbf{mAP$^1$} & \textbf{Seg.} & \textbf{Class.} & \textbf{$\alpha^2=0.3$} \\
    \ChangeRT{1pt}
    
    Rich Features \cite{gupta2014corr}          & NYUDv2  & & & & & 37.3 & &        \\
    \hline
    \RGBD DPM~\cite{gupta2014corr}              & NYUDv2  & & & & & 23.9 & &         \\
    \hline
    RGB R-CNN\cite{gupta2014corr}               & NYUDv2  & & & & & 22.5 & &         \\
    \hline
    Amodal 3D \cite{Deng_2017_CVPR}             & NYUDv2  & 40.9    & & & & & &      \\
    \hline
    \textbf{DSS \cite{Song_2016_CVPR}}          & NYUDv2  & \textbf{36.3}    & & & & & &     \\
    \hline
    VoxNet \cite{maturana2015voxnet}            & NYUDv2  & & & & & & 71.0  &      \\
    \hline
    ShapeNet \cite{wu2015shapenets}             & NYUDv2  & & & & & & 58.0   &     \\
    
    \ChangeRT{1pt}
    
    \textbf{Frustrum PointNets \cite{Qi_2018_CVPR}}     & SUN \RGBD$^2$   & \textbf{54.0}  & & & & & &          \\
    \hline
    \textbf{DSS \cite{Song_2016_CVPR} }                 & SUN \RGBD$^2$   & \textbf{42.1}    & & & & & &        \\
    \hline
    COG \cite{ren2016cogs}                              & SUN \RGBD$^2$   & 47.6  & & & & & &        \\
    \hline
    2D-driven \cite{Lahoud_2017_ICCV}                   & SUN \RGBD$^2$   & 45.1   & & & & & &       \\
    \hline
    Faster R-CNN \cite{ren2015fasterrcnn}               & SUN \RGBD   & & 50.8    & & & & &       \\
    \hline
    \textbf{Unified CNN \cite{sun2018unified}}          & SUN \RGBD   & & \textbf{52.4}  & & & & &           \\
    
    \ChangeRT{1pt}
    
    \textbf{Frustrum PointNets \cite{Qi_2018_CVPR}}     & KITTI (valid)$^3$ & & & \textbf{63.7} & & & & \\
    \hline
    MV3D \cite{chen2017multi}                           & KITTI (valid)$^3$  & & & 55.1  & & & &       \\
    \hline
    MV3D \cite{chen2017multi}                           & KITTI (test)$^3$ & & & & 79.8  & & &       \\
    \hline
    3D FCN \cite{li2016fcnpointcloud}                   & KITTI (test)$^3$ & & & & 68.3 & & &         \\
    \hline
    Vote3Deep \cite{engelcke2017vote3deep}              & KITTI (test)$^3$ & & & & 63.2 & & &        \\
    \hline
    Vote3D \cite{wang2015votingfv}                      & KITTI (test)$^3$ & & & & 42.6  & & &      \\
    \hline
    VeloFCN \cite{li2016vehicle}                        & KITTI (test)$^3$ & & & & 46.9 & & &        \\
    
    \ChangeRT{1pt} 
    
    Complement-Aware \cite{Chen_2018_CVPR}              & NLPR   & & & & & & & 0.85        \\
    \hline
    Salient Deep Fusion \cite{qu2017salientfuse}        & NLPR   & & & & & & & 0.78        \\
    \hline
    LMH \cite{Peng_2014_ECCV}                           & NLPR   & & & & & & & 0.65        \\
    \hline
    GP \cite{ren2015globpriors}                         & NLPR   & & & & & & & 0.72        \\
    \hline
    ACSD \cite{ju2014salientanisotropic}                & NLPR   & & & & & & & 0.54        \\
    
    \ChangeRT{1pt} 
    
    PointNet \cite{qi2017pointnet}                      & ModelNet40   & & & & & & 89.2   &      \\
    \hline
    PointNet++ \cite{qi2017pointnet++}                  & ModelNet40   & & & & & & 91.9 &        \\
    \hline
    VoxNet \cite{qi2017pointnet++}                      & ModelNet40   & & & & & & 85.9 &         \\
    \hline
    3D ShapeNets \cite{Song_2014_ShapeNets}             & ModelNet40   & & & & & & 84.7  &        \\
    \hline
    Subvolumes \cite{qi2016volmulticnn}                 & ModelNet40   & & & & & & 89.2   &       \\
    \hline
    VoxNet \cite{maturana2015voxnet}                    & ModelNet40   & & & & & & 83.0  &      \\
    \hline
    ShapeNet \cite{wu2015shapenets}                     & ModelNet40   & & & & & & 77.0   &     \\
    
    \ChangeRT{1pt} 
    
    Faster R-CNN~\cite{ren2015fasterrcnn}$^{4,5}$       & PASCAL 2012$^6$    & & & &  76.4  & & &      \\
    \hline
    ION~\cite{han2018dlodsurvey}                        & PASCAL 2012$^6$    & & & &  74.6   & & &     \\
    \hline
    SSD512~\cite{han2018dlodsurvey}                     & PASCAL 2012$^6$    & & & &  76.8  & & &      \\
    \hline
    Short Conns \cite{hou2016shortconns}                & PASCAL 2010$^6$   & & & & & & & 0.82         \\
    
    \ChangeRT{1pt} 
    
    MMDL \cite{eitel2015corr}                           & Washington & & & & & &  $91.3$ &         \\
    \hline
    STEM-CaRFs \cite{asif2017cascforest}                & Washington & & & & & &  $88.1$  &       \\
    \hline
    CNN Features \cite{schwarz2015pretrain}             & Washington & & & & & &  $89.4$   &      \\
    
    \ChangeRT{1pt} 
    
    VoxNet \cite{maturana2015voxnet}                    & ModelNet10    & & & & & & 92.0    &    \\
    \hline
    ShapeNet \cite{wu2015shapenets}                     & ModelNet10    & & & & & & 84.0     &   \\
    
    \ChangeRT{1pt} 
    
    Faster R-CNN on FPN \cite{lin2017feature}               & COCO test-dev & & & & 36.2  & & &       \\
    \hline
    AttractioNet \cite{gidaris2016attractio}                & COCO test-dev & & & & 35.7  & & &       \\
    
    \ChangeRT{1pt}
    
    Complement-Aware \cite{Chen_2018_CVPR}                  & NJUDS  & & & & & & & 0.86         \\
    \hline
    Short Conns \cite{hou2016shortconns}                    & MSRA-B & & & & & & & 0.92         \\
    \hline
    MMDL \cite{eitel2015corr}                               & Scenes & & & & & & $82.1$  &        \\
    \ChangeRT{1pt}
    \end{tabular}
    
    $^1$ Values in this column use unspecified IoU thresholds. $^2$ Using only $10$ categories from the SUN \RGBD dataset.  $^3$ Using the hard, cars subset of the KITTI dataset. $^3$ Using the hard, cars subset of the KITTI dataset. $^4$ Results taken from \cite{han2018dlodsurvey}. $^5$ Uses ResNet-101. $^6$ Using the PASCAL Visual Object Classes (VOC) evaluation.
    
\end{table*}

\subsection{Discussion and performance comparison}
\label{sec:comparison}

Tables~\ref{tab:tradperf} and ~\ref{tab:dlperf} summarize the performance,   on various datasets and using various performance metrics\index{performance metric},  of some traditional and deep learning-based \RGBD object detection pipelines. Below, we discuss the pipelines whose performances are highlighted in bold in Table~\ref{tab:dlperf}, with a focus on analyzing the operational steps which allow them to offer increased performance.

The Deep Sliding Shapes\index{deep sliding shape} model for amodal 3D object detection\index{object detection!amodal object detection} in \RGBD images~\cite{Song_2016_CVPR} extends its predecessor \cite{song2014sliding}, which used hand-crafted features\index{feature!hand-crafted} and SVMs\index{SVM}. The network begins by replacing the 3D exhaustive search with a 3D multi-scale RPN,  which produces 3D RoIs. Non-maximum suppression with an IoU constraint of less than  $0.35$ is enforced on the RPN output to reduce the number of RoIs. Each RoI is then projected to 2D and fed to a VGG-19-based \index{VGG} deep feature extractor~\cite{karen2014vgg}, which produces the class labels as well as the 3D bounding boxes of the detected objects. 


Sun \etal~\cite{sun2018unified} proposed an   object detection framework for a mobile manipulation platform. The framework is composed of an RPN, an ORN, and a Scene Recognition Network (SRN).  Its main feature is that the convolutional operations\index{convolutional operation} of the three modules  are shared, subsequently reducing  the computational cost.  The ORN, which  achieved a mean average precision (mAP)\index{mean average precision} of $52.4\%$ on the SUN \RGBD dataset,  outperformed  Faster R-CNN~\cite{ren2015fasterrcnn}, \RGBD RCNN~\cite{gupta2014corr}, and DPM~\cite{felzenszwalb2010dpm}.

Another example is the object detection pipeline of Qi \etal~\cite{Qi_2018_CVPR}, which produces the full extents of an object's bounding box in 3D from \RGBD images by using four sub-networks, namely: 
\begin{itemize}
    \item \textbf{A joint 2D RPN/ORN. } It generates 2D region proposals from the  RGB image,  and classifies them  into one of the $n_c$ object categories.

    \item \textbf{A PointNet-based network. } It performs 3D instance segmentation of the point clouds within 3D frustums extended from the proposed regions.

    \item \textbf{A light-weight regression PointNet (T-Net). } It estimates the true center of the complete object and then transforms the coordinates such that the predicted center becomes the origin.

    \item \textbf{A box estimation network. } It predicts, for each object, its amodal bounding box for the entire object even if parts of it are occluded.
\end{itemize}

\noi The approach simultaneously trains the 3D instance segmentation PointNet\index{PointNet}, the T-Net\index{T-Net},  and the amodal box estimation PointNet, using a loss function\index{loss functon} that is defined as a weighted sum of the losses of the individual subnetworks. Note that Qi \etal~\cite{Qi_2018_CVPR}'s  network architecture is similar to the architecture of the object classification network of~\cite{qi2017pointnet,qi2017pointnet++} but it outputs the  object class scores as well as the detected object's bounding box.  The work also shares some similarities to \cite{Lahoud_2017_ICCV}, which used hand-crafted features.


\section{Summary and perspectives}
\label{sec:summary}

In this chapter, we have reviewed some of the  recent advances in object detection from \RGBD images. Initially we focused on traditional methods, which are based on hand-crafted features combined with machine learning techniques. We then shifted our attention to more recent techniques, which are based on deep learning networks. In terms of performance, deep learning-based techniques significantly outperform traditional methods. However, these methods require large datasets for efficient training. We expect that in the near future, this will become less of an issue since \RGBD sensors are becoming cheaper and annotated \RGBD datasets will thus become widely available.

Although they achieve remarkable performance compared to traditional methods, deep learning techniques are still in their infancy. For instance, amodal object detection, \ie estimating the entire 3D bounding box of an object, especially when parts of the object are occluded, still remains challenging especially in highly cluttered scenes.  Self-occluding objects also challenge deep learning-based pipelines, especially when dealing with dynamic objects that deform in a non-rigid way~\cite{laga2017numerical,jermyn2017elastic}. Similarly, object detection performance for objects at various scales, particularly at small scales, is still relatively low. In fact, the performance comparison of  Table~\ref{tab:dlperf} does not consider the robustness of the methods to scale variation.


Existing  techniques focus mainly on the detection of the bounding boxes of the objects of interest. However, many situations, \eg robust grasping, image editing, and accurate robot navigation, require the accurate detection of object boundaries. Several works have attempted to achieve this using, for example, template matching. This, however, remains an open problem.

Another important avenue for future research is how to incorporate spatial relationships\index{spatial relationship} and relationships between semantic classes in deep learning-based \RGBD object detection pipelines. These relationships are important cues for recognition and it has been already shown in many papers that they can significantly boost the performance of traditional techniques~\cite{laga2013geometry}. Yet, this knowledge is not efficiently exploited in deep learning techniques. 


Finally, there are many aspects of object detection from \RGBD images that have not been covered in this chapter. Examples include saliency detection~\cite{lei2017salientopt, qu2017salientfuse, song2016salientbagclust,Chen_2018_CVPR}, which  aims to detect salient regions in an \RGBD image. Additionally, we have focused in this chapter on generic objects in indoor and outdoor scenes. There is, however, a rich literature on specialized detectors which focus on specific classes of objects, \eg  humans and human body parts such as faces and hands.  


\begin{acknowledgement}
This work is supported by ARC DP $150100294$ and ARC DP $150104251$.
\end{acknowledgement}

\addcontentsline{toc}{section}{Appendix}


\bibliographystyle{spmpsci}
\bibliography{Bennamoun}

\begin{thebibliography}{10}
\providecommand{\url}[1]{{#1}}
\providecommand{\urlprefix}{URL }
\expandafter\ifx\csname urlstyle\endcsname\relax
  \providecommand{\doi}[1]{DOI~\discretionary{}{}{}#1}\else
  \providecommand{\doi}{DOI~\discretionary{}{}{}\begingroup
  \urlstyle{rm}\Url}\fi

\bibitem{cornellgraspingdataset}
Cornell grasping dataset.
\newblock
  \urlprefix\url{http://pr.cs.cornell.edu/grasping/rect\_data/data.php}.
\newblock Accessed: 2018-12-13

\bibitem{achanta2012slic}
Achanta, R., Shaji, A., Smith, K., Lucchi, A., Fua, P., S{\"u}sstrunk, S.: Slic
  superpixels compared to state-of-the-art superpixel methods.
\newblock IEEE Transactions on Pattern Analysis and Machine Intelligence
  \textbf{34}(11), 2274--2282 (2012).
\newblock \doi{10.1109/TPAMI.2012.120}

\bibitem{alexandre2014objrec}
Alexandre, L.A.: 3d object recognition using convolutional neural networks with
  transfer learning between input channels.
\newblock In: IAS (2014)

\bibitem{alexe2014whatisanobject}
Alexe, B., Deselaers, T., Ferrari, V.: What is an object?
\newblock In: 2010 IEEE Computer Society Conference on Computer Vision and
  Pattern Recognition, pp. 73--80 (2010).
\newblock \doi{10.1109/CVPR.2010.5540226}

\bibitem{arbel2014mcg}
Arbel\'{a}ez, P., Pont-Tuset, J., Barron, J., Marques, F., Malik, J.:
  Multiscale combinatorial grouping.
\newblock In: Computer Vision and Pattern Recognition (2014)

\bibitem{asif2017cascforest}
Asif, U., Bennamoun, M., Sohel, F.A.: Rgb-d object recognition and grasp
  detection using hierarchical cascaded forests.
\newblock IEEE Transactions on Robotics \textbf{33}(3), 547--564 (2017).
\newblock \doi{10.1109/TRO.2016.2638453}

\bibitem{barrow1977chamfer}
Barrow, H.G., Tenenbaum, J.M., Bolles, R.C., Wolf, H.C.: Parametric
  correspondence and chamfer matching: Two new techniques for image matching.
\newblock In: Proceedings of the 5th International Joint Conference on
  Artificial Intelligence - Volume 2, IJCAI'77, pp. 659--663. Morgan Kaufmann
  Publishers Inc., San Francisco, CA, USA (1977).
\newblock \urlprefix\url{http://dl.acm.org/citation.cfm?id=1622943.1622971}

\bibitem{bleyer2012extracting}
Bleyer, M., Rhemann, C., Rother, C.: Extracting 3d scene-consistent object
  proposals and depth from stereo images.
\newblock In: European Conference on Computer Vision, pp. 467--481. Springer
  (2012)

\bibitem{bo2014learning}
Bo, L., Ren, X., Fox, D.: Learning hierarchical sparse features for rgb-(d)
  object recognition.
\newblock The International Journal of Robotics Research \textbf{33}(4),
  581--599 (2014)

\bibitem{buch2009hog3d}
Buch, N.E., Orwell, J., Velastin, S.A.: 3d extended histogram of oriented
  gradients (3dhog) for classification of road users in urban scenes.
\newblock In: BMVC (2009)

\bibitem{Chen_2018_CVPR}
Chen, H., Li, Y.: Progressively complementarity-aware fusion network for rgb-d
  salient object detection.
\newblock In: The IEEE Conference on Computer Vision and Pattern Recognition
  (CVPR) (2018)

\bibitem{chen2017multi}
Chen, X., Ma, H., Wan, J., Li, B., Xia, T.: Multi-view 3{D} object detection
  network for autonomous driving.
\newblock In: IEEE CVPR, vol.~1, p.~3 (2017)

\bibitem{dalal2005hogdesc}
Dalal, N., Triggs, B.: Histograms of oriented gradients for human detection.
\newblock In: 2005 IEEE Computer Society Conference on Computer Vision and
  Pattern Recognition (CVPR'05), vol.~1, pp. 886--893 vol. 1 (2005).
\newblock \doi{10.1109/CVPR.2005.177}

\bibitem{lowe2004sift}
{David G. Lowe}: Distinctive image features from scale-invariant keypoints.
\newblock International Journal of Computer Vision (IJCV)  (2004)

\bibitem{Deng_2017_CVPR}
Deng, Z., Latecki, L.J.: {Amodal detection of 3D objects: Inferring 3D bounding
  boxes from 2D ones in RGB-depth images}.
\newblock In: Conference on Computer Vision and Pattern Recognition (CVPR),
  vol.~2, p.~2 (2017)

\bibitem{schapire2013adaboost}
E.~Schapire, R.: Explaining AdaBoost, pp. 37--52 (2013).
\newblock \doi{10.1007/978-3-642-41136-6-5}

\bibitem{eitel2015corr}
Eitel, A., Springenberg, J.T., Spinello, L., Riedmiller, M.A., Burgard, W.:
  Multimodal deep learning for robust {RGB-D} object recognition.
\newblock CoRR \textbf{abs/1507.06821} (2015).
\newblock \urlprefix\url{http://arxiv.org/abs/1507.06821}

\bibitem{engelcke2017vote3deep}
Engelcke, M., Rao, D., Wang, D.Z., Tong, C.H., Posner, I.: {Vote3deep: Fast
  object detection in 3D point clouds using efficient convolutional neural
  networks}.
\newblock In: Robotics and Automation (ICRA), 2017 IEEE International
  Conference on, pp. 1355--1361. IEEE (2017)

\bibitem{pascalvocchallenge}
Everingham, M., Van~Gool, L., Williams, C., Winn, J., Zisserman, A.: The pascal
  visual object classes (voc) challenge.
\newblock In: 2010 IEEE Conference on Computer Vision and Pattern Recognition
  (CVPR) (2010)

\bibitem{feifei2004caltech101}
Fei-Fei, L., Fergus, R., Perona, P.: Learning generative visual models from few
  training examples: An incremental bayesian approach tested on 101 object
  categories.
\newblock 2004 Conference on Computer Vision and Pattern Recognition Workshop
  pp. 178--178 (2004)

\bibitem{felzenszwalb2010dpm}
Felzenszwalb, P.F., Girshick, R.B., McAllester, D., Ramanan, D.: Object
  detection with discriminatively trained part-based models.
\newblock IEEE Transactions on Pattern Analysis and Machine Intelligence
  \textbf{32}(9), 1627--1645 (2010)

\bibitem{Feng_2016_CVPR}
Feng, D., Barnes, N., You, S., McCarthy, C.: Local background enclosure for
  rgb-d salient object detection.
\newblock In: 2016 IEEE Conference on Computer Vision and Pattern Recognition
  (CVPR), pp. 2343--2350 (2016).
\newblock \doi{10.1109/CVPR.2016.257}

\bibitem{geiger2012kitti}
Geiger, A., Lenz, P., Urtasun, R.: Are we ready for autonomous driving? the
  kitti vision benchmark suite.
\newblock In: Conference on Computer Vision and Pattern Recognition (CVPR)
  (2012)

\bibitem{getto2015retrieval}
Getto, R., Fellner, D.W.: 3d object retrieval with parametric templates.
\newblock In: Proceedings of the 2015 Eurographics Workshop on 3D Object
  Retrieval, 3DOR '15, pp. 47--54. Eurographics Association, Goslar Germany,
  Germany (2015).
\newblock \doi{10.2312/3dor.20151054}.
\newblock \urlprefix\url{https://doi.org/10.2312/3dor.20151054}

\bibitem{gidaris2016attractio}
Gidaris, S., Komodakis, N.: Attend refine repeat: Active box proposal
  generation via in-out localization.
\newblock CoRR \textbf{abs/1606.04446} (2016).
\newblock \urlprefix\url{http://arxiv.org/abs/1606.04446}

\bibitem{girshick2015fast}
Girshick, R.B.: Fast {R-CNN}.
\newblock CoRR \textbf{abs/1504.08083} (2015).
\newblock \urlprefix\url{http://arxiv.org/abs/1504.08083}

\bibitem{griffin2007caltech256}
Griffin, G., Holub, A., Perona, P.: Caltech-256 object category dataset.
\newblock Tech. Rep. 7694, California Institute of Technology (2007).
\newblock \urlprefix\url{http://authors.library.caltech.edu/7694}

\bibitem{Gupta_2015_CVPR}
Gupta, S., Arbel{\'a}ez, P., Girshick, R., Malik, J.: Aligning 3d models to
  rgb-d images of cluttered scenes.
\newblock In: Proceedings of the IEEE Conference on Computer Vision and Pattern
  Recognition, pp. 4731--4740 (2015)

\bibitem{gupta2014corr}
Gupta, S., Girshick, R.B., Arbelaez, P., Malik, J.: Learning rich features from
  {RGB-D} images for object detection and segmentation.
\newblock CoRR \textbf{abs/1407.5736} (2014).
\newblock \urlprefix\url{http://arxiv.org/abs/1407.5736}

\bibitem{han2018dlodsurvey}
Han, J., Zhang, D., Cheng, G., Liu, N., Xu, D.: Advanced deep-learning
  techniques for salient and category-specific object detection: A survey.
\newblock IEEE Signal Processing Magazine \textbf{35}(1), 84--100 (2018).
\newblock \doi{10.1109/MSP.2017.2749125}

\bibitem{he1990lbp}
He, D., Wang, L.: Texture unit, texture spectrum, and texture analysis.
\newblock IEEE Transactions on Geoscience and Remote Sensing \textbf{28}(4),
  509--512 (1990).
\newblock \doi{10.1109/TGRS.1990.572934}

\bibitem{hou2016shortconns}
Hou, Q., Cheng, M., Hu, X., Borji, A., Tu, Z., Torr, P.H.S.: Deeply supervised
  salient object detection with short connections.
\newblock CoRR \textbf{abs/1611.04849} (2016).
\newblock \urlprefix\url{http://arxiv.org/abs/1611.04849}

\bibitem{jermyn2017elastic}
Jermyn, I.H., Kurtek, S., Laga, H., Srivastava, A.: Elastic shape analysis of
  three-dimensional objects.
\newblock Synthesis Lectures on Computer Vision \textbf{12}(1), 1--185 (2017)

\bibitem{jiang2014finding}
Jiang, H.: Finding approximate convex shapes in rgbd images.
\newblock In: European Conference on Computer Vision, pp. 582--596. Springer
  (2014)

\bibitem{jiang2013linear}
Jiang, H., Xiao, J.: A linear approach to matching cuboids in rgbd images.
\newblock In: Proceedings of the IEEE Conference on Computer Vision and Pattern
  Recognition, pp. 2171--2178 (2013)

\bibitem{ju2014salientanisotropic}
Ju, R., Ge, L., Geng, W., Ren, T., Wu, G.: Depth saliency based on anisotropic
  center-surround difference.
\newblock In: 2014 IEEE International Conference on Image Processing (ICIP),
  pp. 1115--1119 (2014).
\newblock \doi{10.1109/ICIP.2014.7025222}

\bibitem{kendall2017geo}
Kendall, A., Cipolla, R.: Geometric loss functions for camera pose regression
  with deep learning.
\newblock CoRR \textbf{abs/1704.00390} (2017).
\newblock \urlprefix\url{http://arxiv.org/abs/1704.00390}

\bibitem{khan2018cnnguide}
Khan, S., Rahmani, H., Shah, S.A.A., Bennamoun, M.: A Guide to Convolutional
  Neural Networks for Computer Vision.
\newblock Morgan and Claypool Publishers (2018)

\bibitem{khan2015separating}
Khan, S.H., He, X., Bennamoun, M., Sohel, F., Togneri, R.: Separating objects
  and clutter in indoor scenes.
\newblock In: Proceedings of the IEEE Conference on Computer Vision and Pattern
  Recognition, pp. 4603--4611 (2015)

\bibitem{krizhevsky2012imagenetpretrained}
Krizhevsky, A., Sutskever, I., Hinton, G.E.: Imagenet classification with deep
  convolutional neural networks.
\newblock In: Proceedings of the 25th International Conference on Neural
  Information Processing Systems - Volume 1, NIPS'12, pp. 1097--1105. Curran
  Associates Inc., USA (2012).
\newblock \urlprefix\url{http://dl.acm.org/citation.cfm?id=2999134.2999257}

\bibitem{laga20183d}
Laga, H., Guo, Y., Tabia, H., Fisher, R.B., Bennamoun, M.: 3D Shape Analysis:
  Fundamentals, Theory, and Applications.
\newblock John Wiley \& Sons (2018)

\bibitem{laga20193d}
Laga, H., Guo, Y., Tabia, H., Fisher, R.B., Bennamoun, M.: 3D Shape Analysis:
  Fundamentals, Theory, and Applications.
\newblock Wiley (2019)

\bibitem{laga2013geometry}
Laga, H., Mortara, M., Spagnuolo, M.: Geometry and context for semantic
  correspondences and functionality recognition in man-made 3d shapes.
\newblock ACM Transactions on Graphics (TOG) \textbf{32}(5), 150 (2013)

\bibitem{laga2017numerical}
Laga, H., Xie, Q., Jermyn, I.H., Srivastava, A.: Numerical inversion of srnf
  maps for elastic shape analysis of genus-zero surfaces.
\newblock IEEE transactions on pattern analysis and machine intelligence
  \textbf{39}(12), 2451--2464 (2017)

\bibitem{Lahoud_2017_ICCV}
Lahoud, J., Ghanem, B.: {2D-Driven 3D Object Detection in RGB-D Images}.
\newblock In: The IEEE International Conference on Computer Vision (ICCV)
  (2017)

\bibitem{lai2013rgb}
Lai, K., Bo, L., Ren, X., Fox, D.: Rgb-d object recognition: Features,
  algorithms, and a large scale benchmark.
\newblock In: Consumer Depth Cameras for Computer Vision, pp. 167--192.
  Springer (2013)

\bibitem{lei2017salientopt}
Lei, Z., Chai, W., Zhao, S., Song, H., Li, F.: Saliency detection for rgb-d
  images using optimization.
\newblock In: 2017 12th International Conference on Computer Science and
  Education (ICCSE), pp. 440--443 (2017).
\newblock \doi{10.1109/ICCSE.2017.8085532}

\bibitem{li2016fcnpointcloud}
Li, B.: 3d fully convolutional network for vehicle detection in point cloud.
\newblock CoRR \textbf{abs/1611.08069} (2016).
\newblock \urlprefix\url{http://arxiv.org/abs/1611.08069}

\bibitem{li2016vehicle}
Li, B., Zhang, T., Xia, T.: {Vehicle detection from 3D lidar using fully
  convolutional network}.
\newblock arXiv preprint arXiv:1608.07916  (2016)

\bibitem{li2017lightfield}
Li, N., Ye, J., Ji, Y., Ling, H., Yu, J.: Saliency detection on light field.
\newblock IEEE Transactions on Pattern Analysis and Machine Intelligence
  \textbf{39}(8), 1605--1616 (2017).
\newblock \doi{10.1109/TPAMI.2016.2610425}

\bibitem{lin2013holistic}
Lin, D., Fidler, S., Urtasun, R.: {Holistic scene understanding for 3D object
  detection with RGBD cameras}.
\newblock In: Proceedings of the IEEE International Conference on Computer
  Vision, pp. 1417--1424 (2013)

\bibitem{lin2017feature}
Lin, T.Y., Doll{\'a}r, P., Girshick, R.B., He, K., Hariharan, B., Belongie,
  S.J.: Feature pyramid networks for object detection.
\newblock In: CVPR, vol.~1, p.~4 (2017)

\bibitem{long2014fcnseg}
Long, J., Shelhamer, E., Darrell, T.: Fully convolutional networks for semantic
  segmentation.
\newblock CoRR \textbf{abs/1411.4038} (2014).
\newblock \urlprefix\url{http://arxiv.org/abs/1411.4038}

\bibitem{maturana2015cnnarch}
Maturana, D., Scherer, S.: 3d convolutional neural networks for landing zone
  detection from lidar.
\newblock In: 2015 IEEE International Conference on Robotics and Automation
  (ICRA), pp. 3471--3478 (2015).
\newblock \doi{10.1109/ICRA.2015.7139679}

\bibitem{maturana2015voxnet}
Maturana, D., Scherer, S.: {VoxNet: A 3D Convolutional Neural Network for
  real-time object recognition}.
\newblock In: Ieee/rsj International Conference on Intelligent Robots and
  Systems, pp. 922--928 (2015)

\bibitem{menze2015kitti}
Menze, M., Geiger, A.: Object scene flow for autonomous vehicles.
\newblock In: Conference on Computer Vision and Pattern Recognition (CVPR)
  (2015)

\bibitem{nakahara2017multiscalewindow}
Nakahara, H., Yonekawa, H., Sato, S.: An object detector based on multiscale
  sliding window search using a fully pipelined binarized cnn on an fpga.
\newblock In: 2017 International Conference on Field Programmable Technology
  (ICFPT), pp. 168--175 (2017).
\newblock \doi{10.1109/FPT.2017.8280135}

\bibitem{silberman2012nyu}
Nathan~Silberman Derek~Hoiem, P.K., Fergus, R.: Indoor segmentation and support
  inference from rgb-d images.
\newblock In: ECCV (2012)

\bibitem{newcombe2011kf}
Newcombe, R.A., Izadi, S., Hilliges, O., Molyneaux, D., Kim, D., Davison, A.J.,
  Kohi, P., Shotton, J., Hodges, S., Fitzgibbon, A.: Kinectfusion: Real-time
  dense surface mapping and tracking.
\newblock In: 2011 10th IEEE International Symposium on Mixed and Augmented
  Reality, pp. 127--136 (2011).
\newblock \doi{10.1109/ISMAR.2011.6092378}

\bibitem{Peng_2014_ECCV}
Peng, H., Li, B., Xiong, W., Hu, W., Ji, R.: Rgb-d salient object detection: A
  benchmark and algorithms.
\newblock In: ECCV (2014)

\bibitem{pont-tuset2015mcg}
Pont-Tuset, J., Arbel\'{a}ez, P., Barron, J., Marques, F., Malik, J.:
  Multiscale combinatorial grouping for image segmentation and object proposal
  generation.
\newblock In: arXiv:1503.00848 (2015)

\bibitem{Qi_2018_CVPR}
Qi, C.R., Liu, W., Wu, C., Su, H., Guibas, L.J.: Frustum pointnets for 3d
  object detection from rgb-d data.
\newblock In: The IEEE Conference on Computer Vision and Pattern Recognition
  (CVPR) (2018)

\bibitem{qi2017pointnet}
Qi, C.R., Su, H., Mo, K., Guibas, L.J.: Pointnet: Deep learning on point sets
  for 3d classification and segmentation.
\newblock Proc. Computer Vision and Pattern Recognition (CVPR), IEEE
  \textbf{1}(2), 4 (2017)

\bibitem{qi2016volmulticnn}
Qi, C.R., Su, H., Nie{\ss}ner, M., Dai, A., Yan, M., Guibas, L.J.: Volumetric
  and multi-view cnns for object classification on 3d data.
\newblock CoRR \textbf{abs/1604.03265} (2016).
\newblock \urlprefix\url{http://arxiv.org/abs/1604.03265}

\bibitem{qi2017pointnet++}
Qi, C.R., Yi, L., Su, H., Guibas, L.J.: Pointnet++: Deep hierarchical feature
  learning on point sets in a metric space.
\newblock In: Advances in Neural Information Processing Systems, pp. 5099--5108
  (2017)

\bibitem{qu2017salientfuse}
Qu, L., He, S., Zhang, J., Tian, J., Tang, Y., Yang, Q.: Rgb-d salient object
  detection via deep fusion.
\newblock IEEE Transactions on Image Processing \textbf{26}(5), 2274--2285
  (2017).
\newblock \doi{10.1109/TIP.2017.2682981}

\bibitem{ren2015globpriors}
Ren, J., Gong, X., Yu, L., Zhou, W., Yang, M.Y.: Exploiting global priors for
  rgb-d saliency detection.
\newblock In: 2015 IEEE Conference on Computer Vision and Pattern Recognition
  Workshops (CVPRW), pp. 25--32 (2015).
\newblock \doi{10.1109/CVPRW.2015.7301391}

\bibitem{ren2015faster}
Ren, S., He, K., Girshick, R., Sun, J.: Faster r-cnn: Towards real-time object
  detection with region proposal networks.
\newblock In: Advances in neural information processing systems, pp. 91--99
  (2015)

\bibitem{ren2015fasterrcnn}
Ren, S., He, K., Girshick, R.B., Sun, J.: Faster {R-CNN:} towards real-time
  object detection with region proposal networks.
\newblock CoRR \textbf{abs/1506.01497} (2015).
\newblock \urlprefix\url{http://arxiv.org/abs/1506.01497}

\bibitem{ren2016cogs}
Ren, Z., Sudderth, E.B.: Three-dimensional object detection and layout
  prediction using clouds of oriented gradients.
\newblock In: 2016 IEEE Conference on Computer Vision and Pattern Recognition
  (CVPR), pp. 1525--1533 (2016).
\newblock \doi{10.1109/CVPR.2016.169}

\bibitem{rusinkiewicz2001icp}
Rusinkiewicz, S., Levoy, M.: Efficient variants of the icp algorithm.
\newblock In: Proceedings Third International Conference on 3-D Digital Imaging
  and Modeling, pp. 145--152 (2001).
\newblock \doi{10.1109/IM.2001.924423}

\bibitem{russakovsky2015imagenet}
Russakovsky, O., Deng, J., Su, H., Krause, J., Satheesh, S., Ma, S., Huang, Z.,
  Karpathy, A., Khosla, A., Bernstein, M., Berg, A.C., Fei-Fei, L.: {ImageNet
  Large Scale Visual Recognition Challenge}.
\newblock International Journal of Computer Vision (IJCV) \textbf{115}(3),
  211--252 (2015).
\newblock \doi{10.1007/s11263-015-0816-y}

\bibitem{sahin2016houghforest}
Sahin, C., Kouskouridas, R., Kim, T.: Iterative hough forest with histogram of
  control points for 6 dof object registration from depth images.
\newblock CoRR \textbf{abs/1603.02617} (2016).
\newblock \urlprefix\url{http://arxiv.org/abs/1603.02617}

\bibitem{sahin2017partextract}
Sahin, C., Kouskouridas, R., Kim, T.: A learning-based variable size part
  extraction architecture for 6d object pose recovery in depth.
\newblock CoRR \textbf{abs/1701.02166} (2017).
\newblock \urlprefix\url{http://arxiv.org/abs/1701.02166}

\bibitem{schwarz2015pretrain}
Schwarz, M., Schulz, H., Behnke, S.: Rgb-d object recognition and pose
  estimation based on pre-trained convolutional neural network features.
\newblock In: 2015 IEEE International Conference on Robotics and Automation
  (ICRA), pp. 1329--1335 (2015).
\newblock \doi{10.1109/ICRA.2015.7139363}

\bibitem{shotton2013efficient}
Shotton, J., Girshick, R., Fitzgibbon, A., Sharp, T., Cook, M., Finocchio, M.,
  Moore, R., Kohli, P., Criminisi, A., Kipman, A., et~al.: Efficient human pose
  estimation from single depth images.
\newblock IEEE Transactions on Pattern Analysis and Machine Intelligence
  \textbf{35}(12), 2821--2840 (2013)

\bibitem{karen2014vgg}
Simonyan, K., Zisserman, A.: Very deep convolutional networks for large-scale
  image recognition.
\newblock arXiv preprint arXiv:1409.1556  (2014)

\bibitem{song2016salientbagclust}
Song, H., Liu, Z., Xie, Y., Wu, L., Huang, M.: {RGBD co-saliency detection via
  bagging-based clustering}.
\newblock IEEE Signal Processing Letters \textbf{23}(12), 1722--1726 (2016)

\bibitem{song2015sun}
Song, S., Lichtenberg, S.P., Xiao, J.: Sun rgb-d: A rgb-d scene understanding
  benchmark suite.
\newblock In: 2015 IEEE Conference on Computer Vision and Pattern Recognition
  (CVPR), pp. 567--576. IEEE (2015)

\bibitem{song2014sliding}
Song, S., Xiao, J.: Sliding shapes for 3d object detection in depth images.
\newblock In: European conference on computer vision, pp. 634--651. Springer
  (2014)

\bibitem{Song_2016_CVPR}
Song, S., Xiao, J.: {D}eep {S}liding {S}hapes for amodal 3{D} object detection
  in {RGB-D} images.
\newblock In: CVPR (2016)

\bibitem{sun2018unified}
Sun, H., Meng, Z., Tao, P.Y., Ang, M.H.: Scene recognition and object detection
  in a unified convolutional neural network on a mobile manipulator.
\newblock In: 2018 IEEE International Conference on Robotics and Automation
  (ICRA), pp. 1--5 (2018).
\newblock \doi{10.1109/ICRA.2018.8460535}

\bibitem{Wang_2015_ICCV}
Wang, A., Cai, J., Lu, J., Cham, T.J.: {MMSS: Multi-modal sharable and specific
  feature learning for RGB-D object recognition}.
\newblock In: Proceedings of the IEEE International Conference on Computer
  Vision, pp. 1125--1133 (2015)

\bibitem{wang2015votingfv}
Wang, D.Z., Posner, I.: Voting for voting in online point cloud object
  detection.
\newblock In: Robotics: Science and Systems (2015)

\bibitem{ward2019lossfns}
Ward, I.R., Jalwana, M.A.A.K., Bennamoun, M.: Improving image-based
  localization with deep learning: The impact of the loss function
  \textbf{abs/1905.03692} (2019).
\newblock \urlprefix\url{http://arxiv.org/abs/1905.03692}

\bibitem{Song_2014_ShapeNets}
Wu, Z., Song, S., Khosla, A., Tang, X., Xiao, J.: 3d shapenets for 2.5d object
  recognition and next-best-view prediction.
\newblock CoRR  (2014)

\bibitem{wu2015shapenets}
Wu, Z., Song, S., Khosla, A., Yu, F., Zhang, L., Tang, X., Xiao, J.: 3d
  shapenets: A deep representation for volumetric shapes.
\newblock In: 2015 IEEE Conference on Computer Vision and Pattern Recognition
  (CVPR), pp. 1912--1920 (2015).
\newblock \doi{10.1109/CVPR.2015.7298801}

\bibitem{xiang2016objectnet3d}
Xiang, Y., Kim, W., Chen, W., Ji, J., Choy, C., Su, H., Mottaghi, R., Guibas,
  L., Savarese, S.: {ObjectNet3D: A large scale database for 3D object
  recognition}.
\newblock In: European Conference on Computer Vision, pp. 160--176. Springer
  (2016)

\bibitem{xiang2014beyond}
Xiang, Y., Mottaghi, R., Savarese, S.: Beyond pascal: A benchmark for 3d object
  detection in the wild.
\newblock In: Applications of Computer Vision (WACV), 2014 IEEE Winter
  Conference on, pp. 75--82. IEEE (2014)

\bibitem{xiao2012sun}
Xiao, J., Hays, J., Ehinger, K.A., Oliva, A., Torralba, A.: Sun database:
  Large-scale scene recognition from abbey to zoo.
\newblock In: 2010 IEEE Computer Society Conference on Computer Vision and
  Pattern Recognition, pp. 3485--3492 (2010).
\newblock \doi{10.1109/CVPR.2010.5539970}

\end{thebibliography}

\backmatter


\end{document}